\documentclass{article}
    \PassOptionsToPackage{numbers, compress}{natbib}

\usepackage[preprint]{neurips_2021}
\usepackage[utf8]{inputenc} 
\usepackage[T1]{fontenc}    
\usepackage{hyperref}       
\usepackage{url}            
\usepackage{booktabs}       
\usepackage{amsfonts}       
\usepackage{nicefrac}       
\usepackage{microtype}      
\usepackage{textcomp}
\usepackage[normalem]{ulem}
\usepackage{authblk}

\newcommand{\BS}{\text{BERT}_{\text{SMALL}}}
\newcommand{\BB}{\text{BERT}_{\text{BASE}}}
\newcommand{\BL}{\text{BERT}_{\text{LARGE}}}
\newcommand{\cov}{\mC}
\newcommand{\qcov}{\mQ}

\usepackage{url}

\usepackage{graphicx} 
\usepackage{subfigure} 
\usepackage{multirow}
\usepackage[dvipsnames]{xcolor}
\usepackage{mathtools}
\usepackage{relsize}
\usepackage{nicefrac}
\usepackage{xspace}
\usepackage{amsmath,amssymb,enumerate}
\usepackage{amsthm,cancel}
\usepackage{natbib}
\usepackage{enumitem}
\usepackage{dsfont}

\newtheorem*{lemma*}{Lemma}
\newtheorem*{theorem*}{Theorem}
\newtheorem*{assumption*}{Assumption}
\newtheorem*{corollary*}{Corollary}
\newtheorem*{remark*}{Remark}
\newtheorem*{definition*}{Definition}

\usepackage{hyperref}
\usepackage{url}

\usepackage[utf8]{inputenc} 
\usepackage[T1]{fontenc}    
\usepackage{url}            
\usepackage{booktabs}       
\usepackage{amsfonts}       
\usepackage{nicefrac}       
\usepackage{microtype}      
\usepackage{algorithm}
\usepackage{algorithmic}

\usepackage{amsmath,amsfonts,bm}

\def\1{\bm{1}}

\def\mA{{\bm{A}}}

\def\mC{{\bm{C}}}

\def\mQ{{\bm{Q}}}

\def\mV{{\bm{V}}}
\def\mW{{\bm{W}}}
\def\mX{{\bm{X}}}

\def\mZ{{\bm{Z}}}

\DeclareMathAlphabet{\mathsfit}{\encodingdefault}{\sfdefault}{m}{sl}
\SetMathAlphabet{\mathsfit}{bold}{\encodingdefault}{\sfdefault}{bx}{n}

\def\gX{{\mathcal{X}}}

\newcommand{\E}{\mathbb{E}}

\newcommand{\R}{\mathbb{R}}

\DeclareMathOperator{\Tr}{Tr}

\title{Eigen Analysis of Self-Attention and its Reconstruction from Partial Computation}

\author{Srinadh Bhojanapalli\thanks{Authors are ordered alphabetically. Corresponding email: bsrinadh@google.com}, Ayan Chakrabarti, Himanshu Jain, \\Sanjiv Kumar, Michal Lukasik, Andreas Veit}
\affil{Google Research, New York}

\begin{document}
\maketitle

\newcommand{\insertFigBLGlobalEigen}{
\begin{figure*}[!t]
    \centering
    \subfigure{
    \includegraphics[scale=0.3]{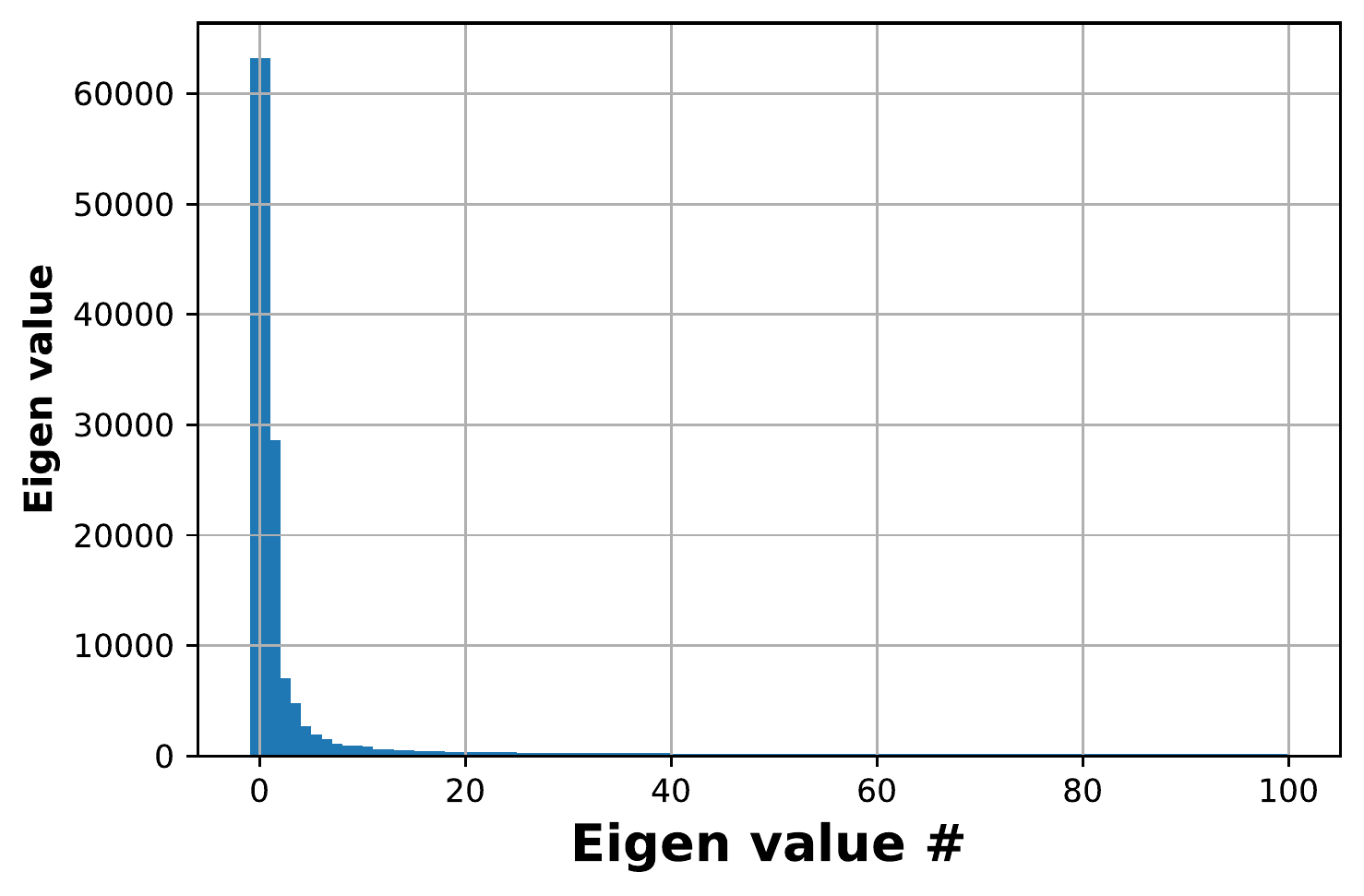}%
    } 
    \hfill
    \subfigure{
    \includegraphics[scale=0.3]{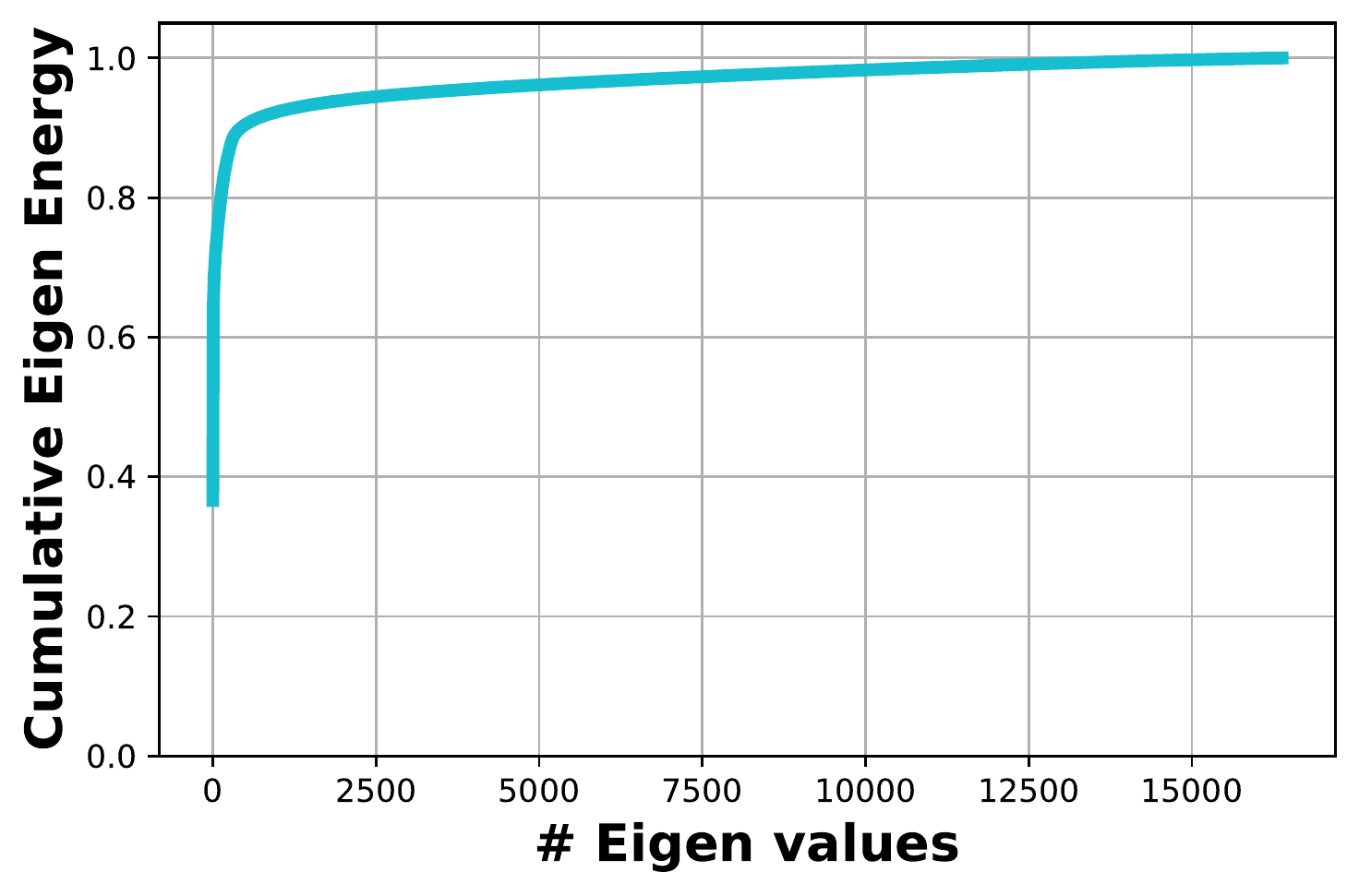}%
    }
    \hfill
    \subfigure{
    \includegraphics[scale=0.3]{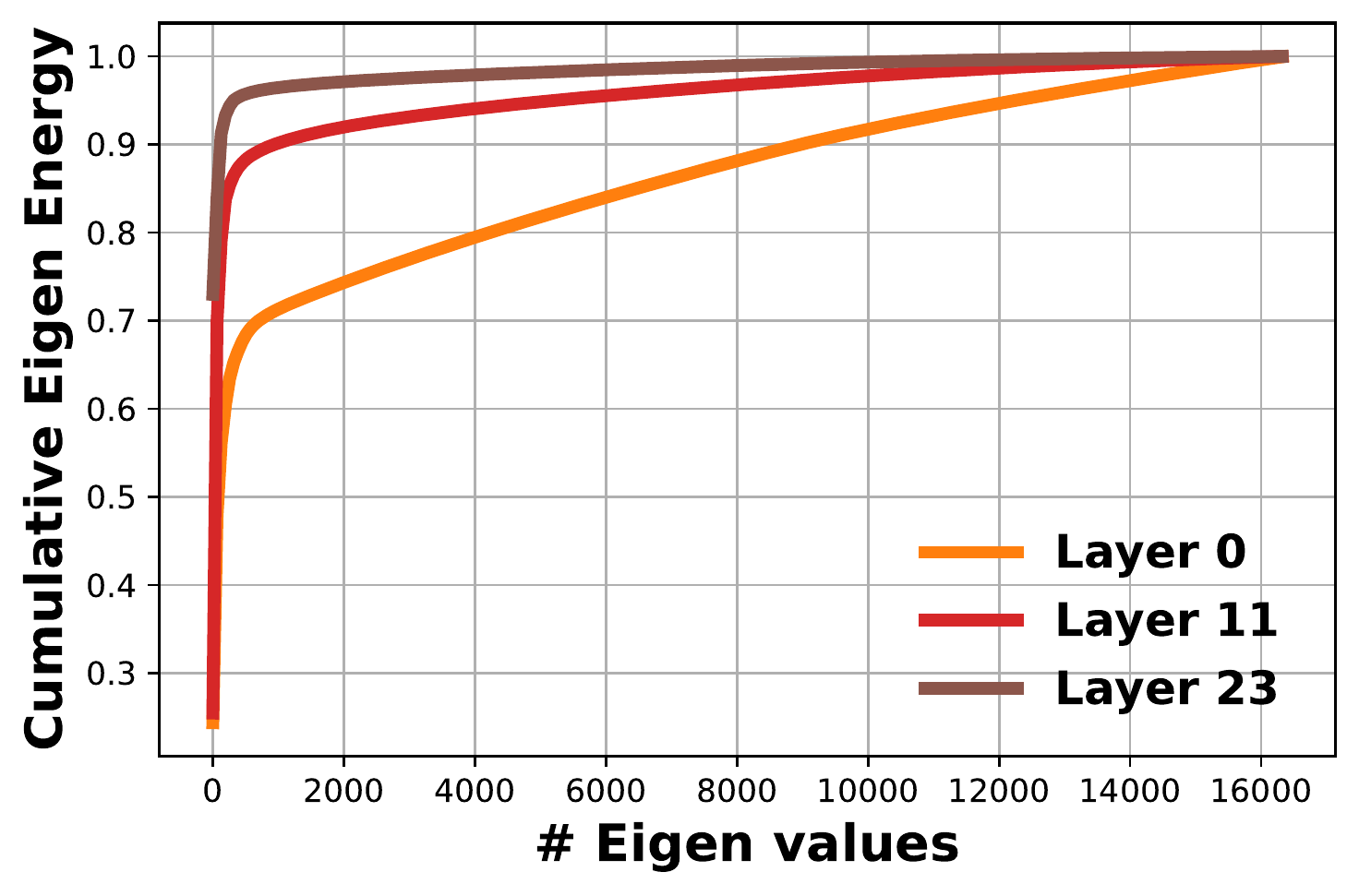}%
    }
\caption{\textbf{Eigen values of $\cov_a$}. Left: Barplot of the top 100 eigen values of the covariance matrix $\cov_a$ of attention scores aggregated over the entire network of a $\BL$ model. Middle: Cumulative sum of eigen values of $\cov_a$. Both show that $\cov_a$ is approximately low rank with top 125 eigen vectors capturing > $80\%$ of the energy. Right: Cumulative sum of eigen values of attention scores covariance matrix $\cov_a^l$ for different layers of a $\BL$ model. We notice that later layers in the network have smaller rank. }
    \label{fig:eigen_bl_global}
\end{figure*}
}

\newcommand{\insertFigBBGlobalEigen}{
\begin{figure*}[!t]
    \centering
    \subfigure{
    \includegraphics[scale=0.3]{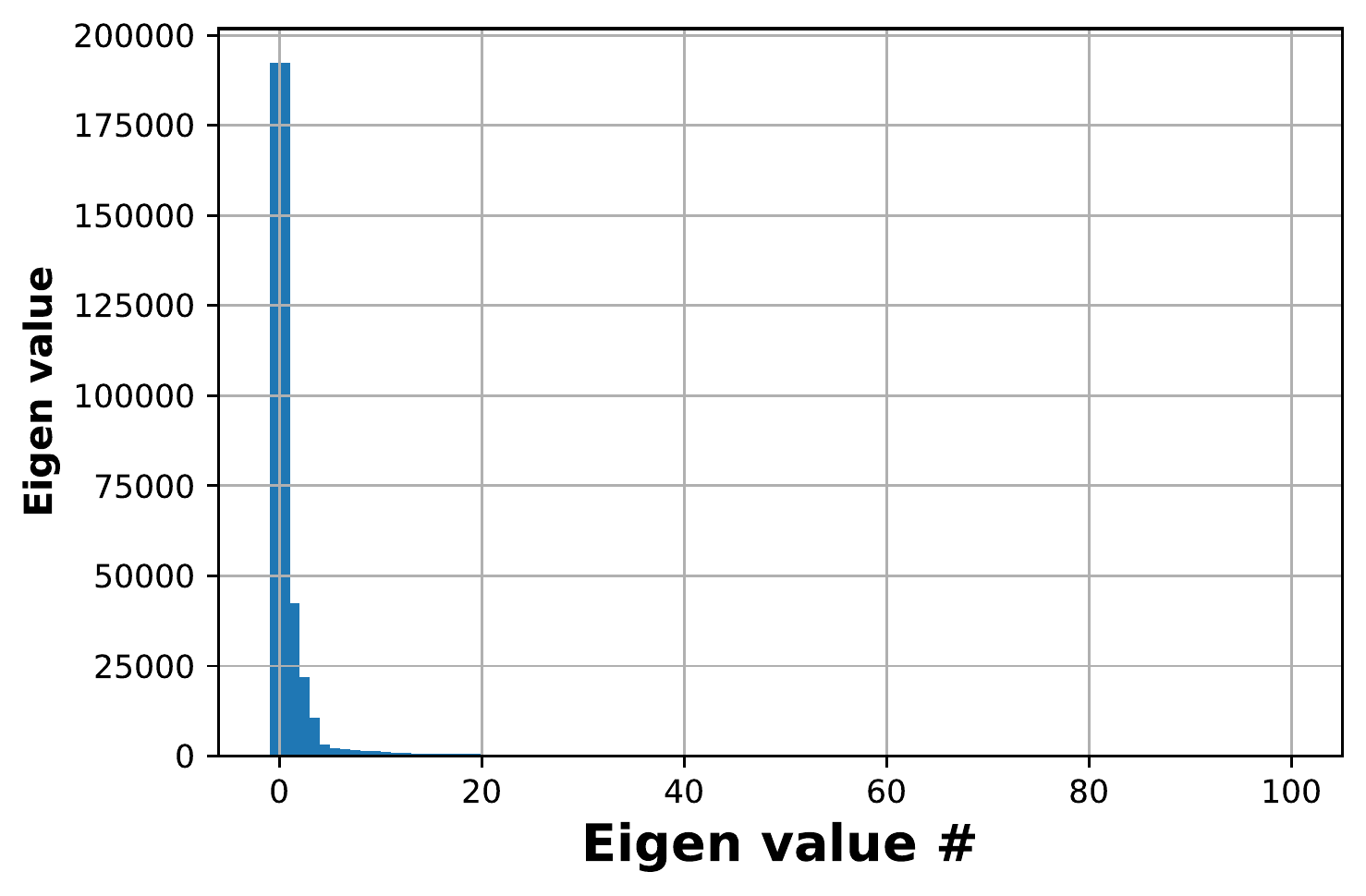}%
    } 
    \hfill
    \subfigure{
    \includegraphics[scale=0.3]{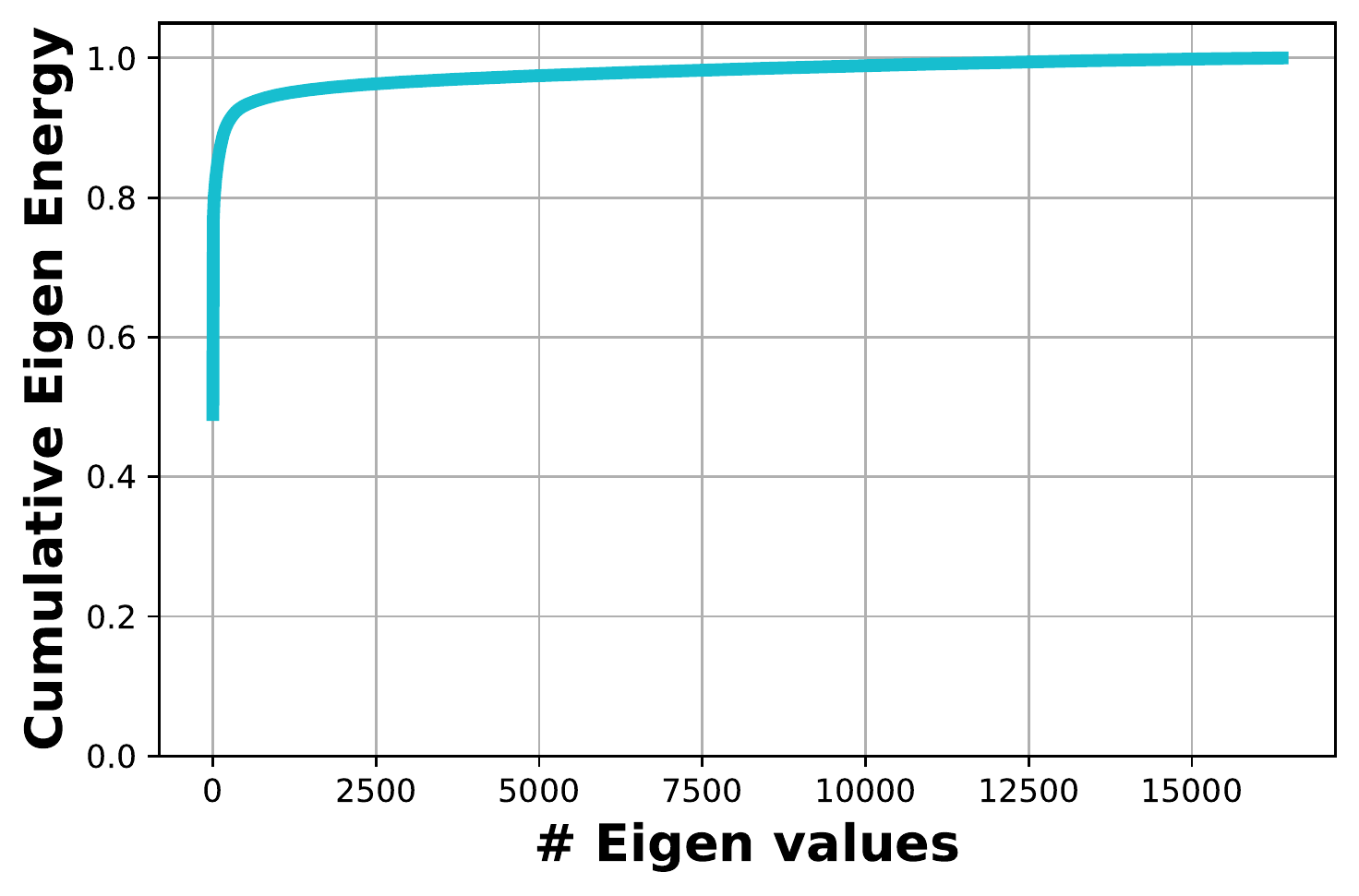}%
    }
    \hfill
    \subfigure{
    \includegraphics[scale=0.3]{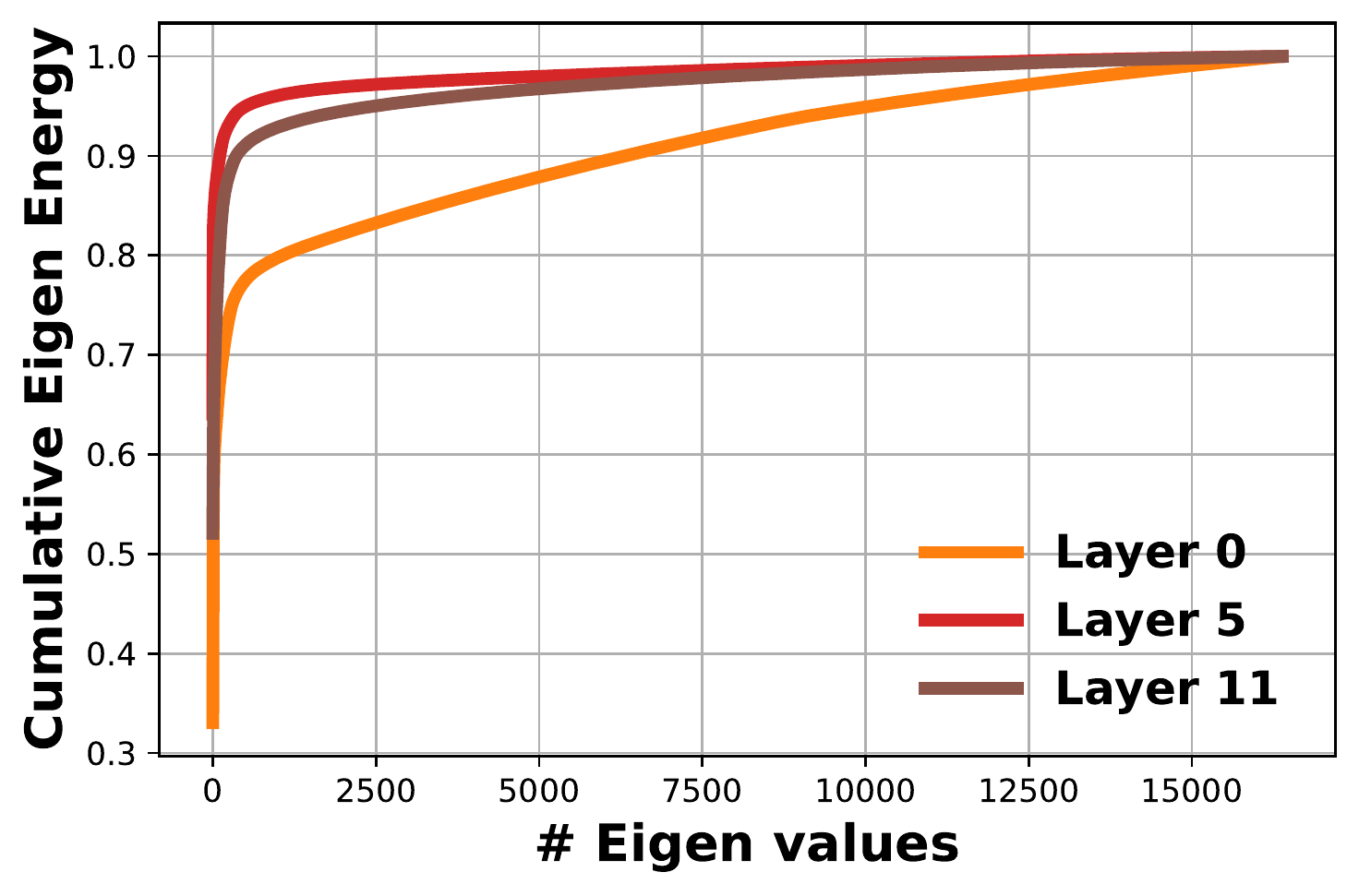}%
    }
\caption{\textbf{Eigen values of $\cov_a$}. Left: Barplot of the top 100 eigen values of the covariance matrix $\cov_a$ of attention scores aggregated over the entire network of a $\BB$ model. Middle: Cumulative sum of eigen values of $\cov_a$. Both show that $\cov_a$ is approximately low rank with top 200 ($1.2\%$) eigen values capturing > $90\%$ of the energy.  Right: Cumulative sum of eigen values of attention scores covariance matrix $\cov_a^l$ for different layers of a $\BB$ model. We notice later layers in the network have smaller rank.}
    \label{fig:eigen_bb_global}
    \vspace{-0.05in}
\end{figure*}
}

\newcommand{\insertFigBLEigenAll}{
\begin{figure*}[!t]
        \centering
    \subfigure{
    \includegraphics[scale=0.3]{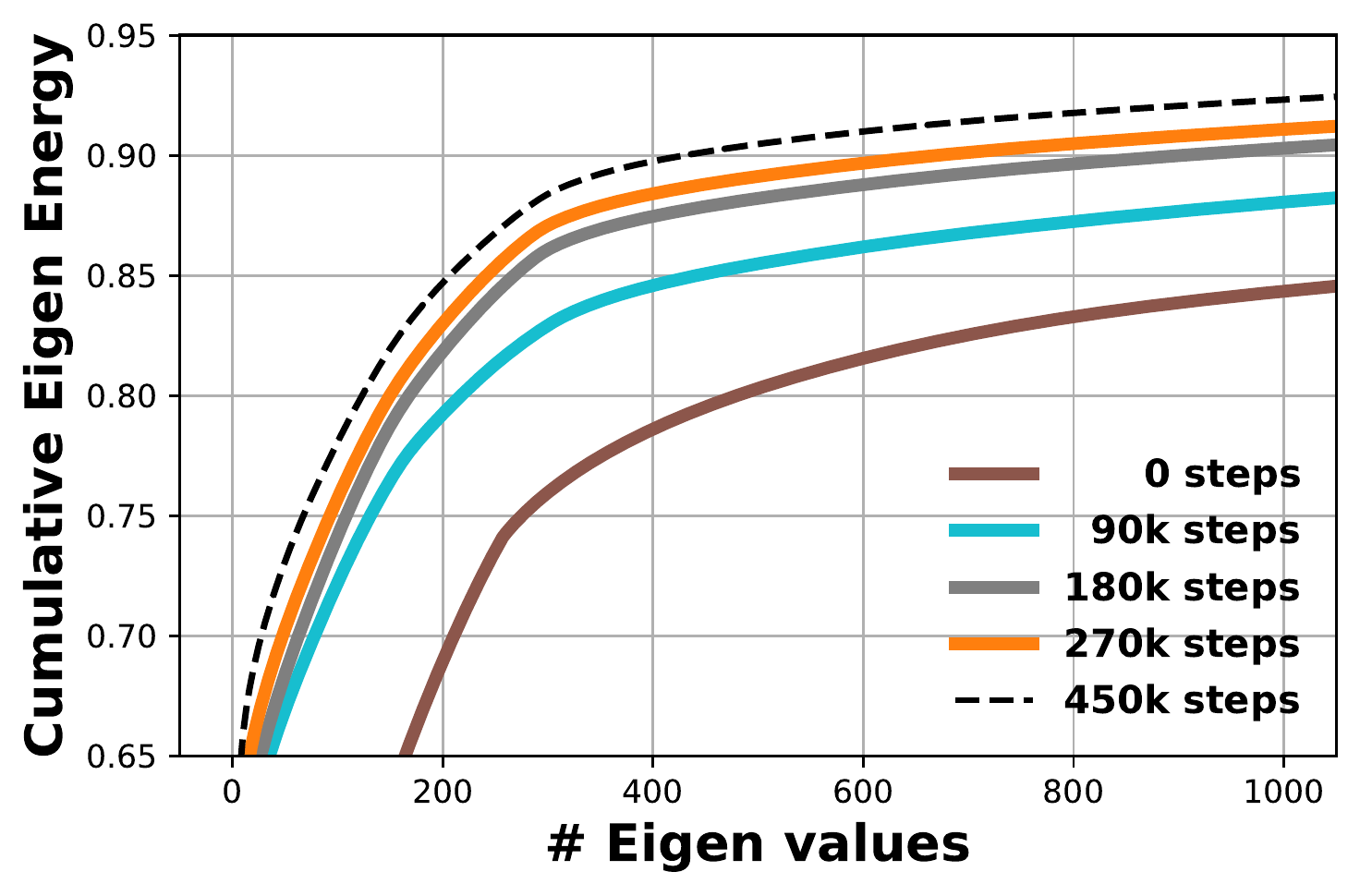}%
    }
    \hfill
    \subfigure{
    \includegraphics[scale=0.3]{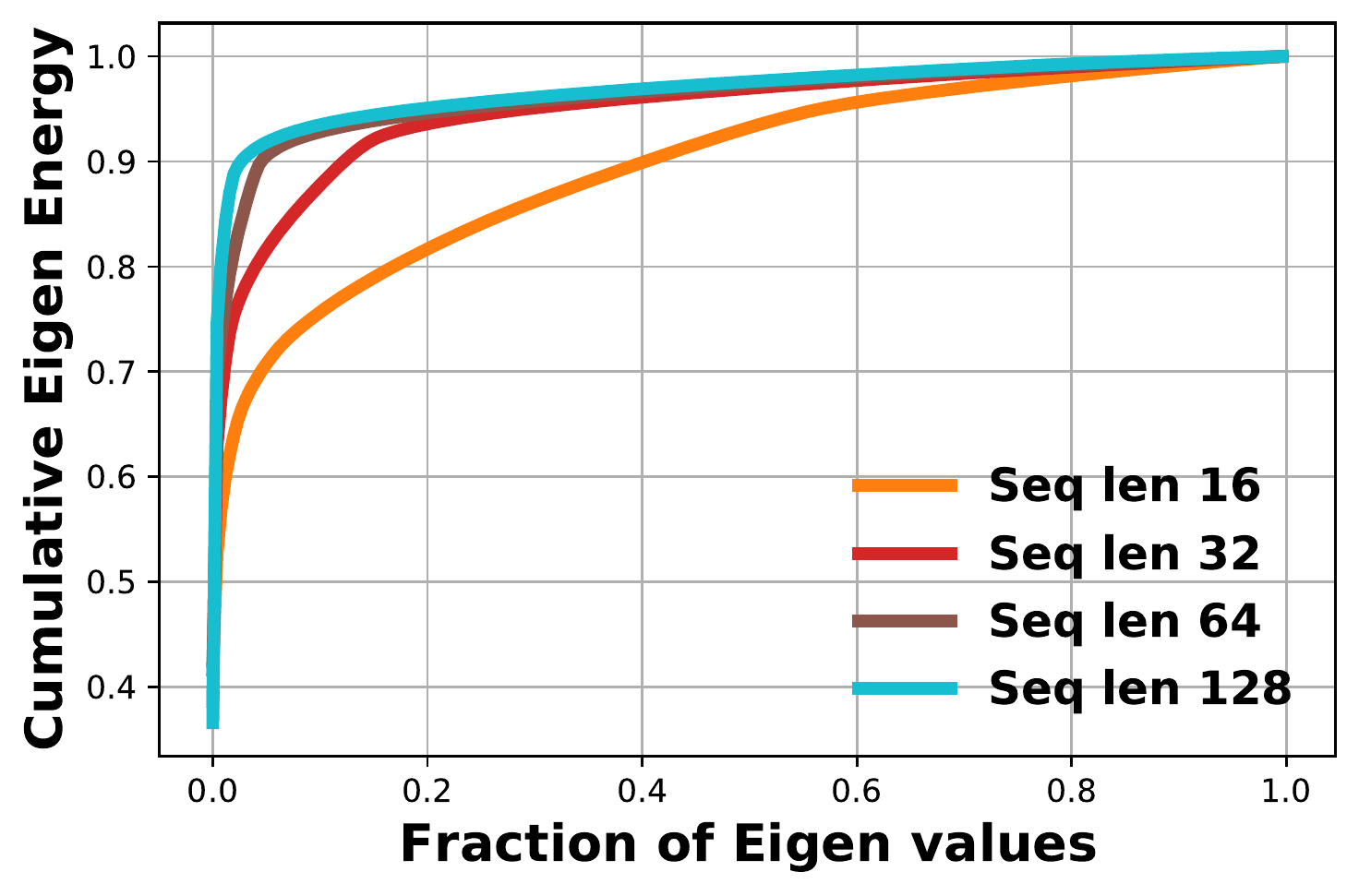}%
    } 
    \hfill
    \subfigure{
    \includegraphics[scale=0.3]{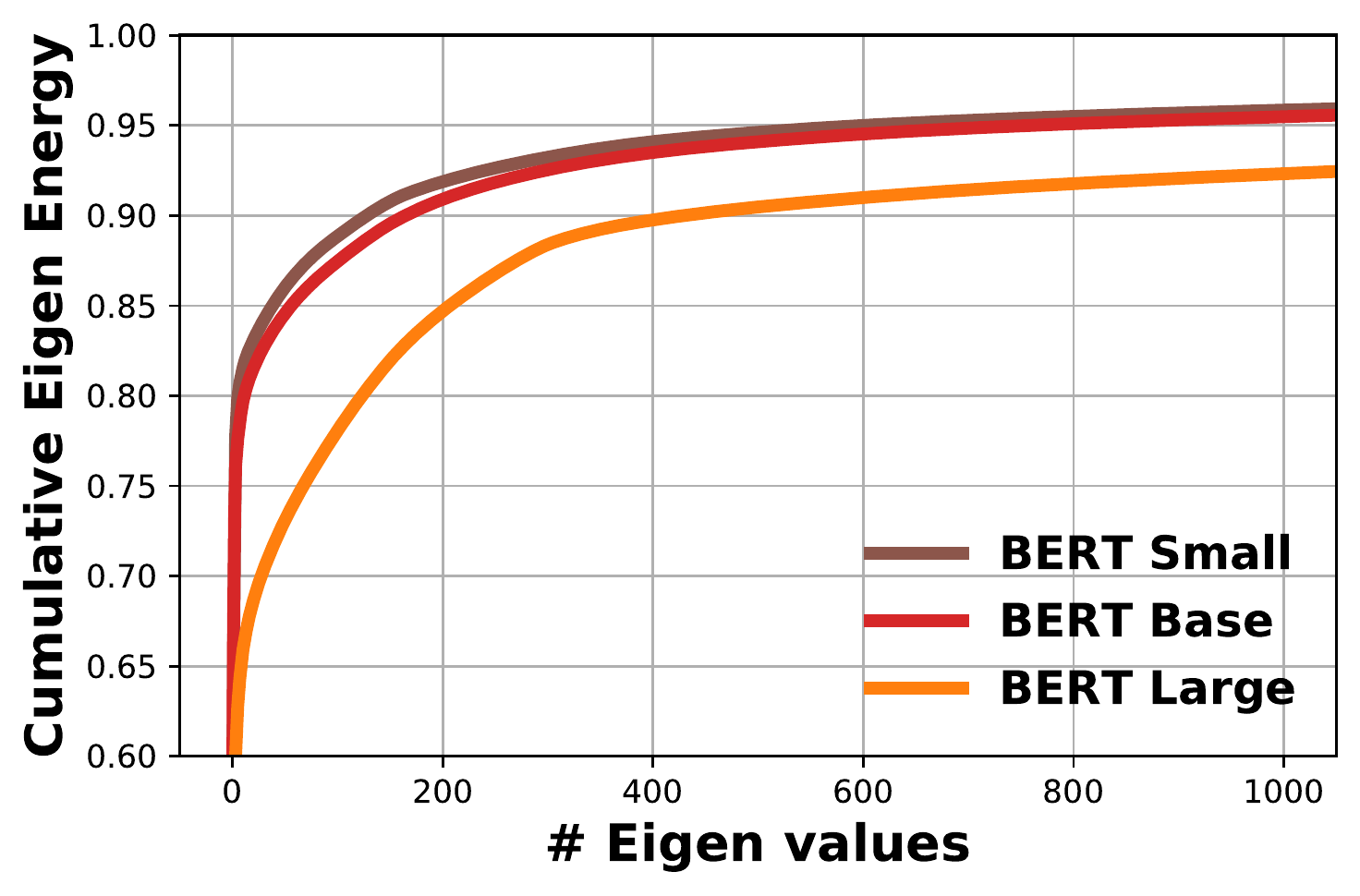}%
    } 
    \caption{\textbf{Eigen values of $\cov_a$}. Cumulative sum of eigen values of attention scores covariance matrix $\cov_a$ of a $\BL$ model  - Left: after varying number of training steps. We notice that the rank slightly decreases throughout training with a large reduction in the beginning. Middle: for different sequence length inputs. Note that x-axis here denotes the fraction of eigen values. Right: for varying model sizes. Note that rank slightly increases with model size. 
    }
    \label{fig:eigen_bl_ablation}
\end{figure*}
}

\newcommand{\insertFigBBEigenAll}{
\begin{figure*}[!t]
        \centering
    \subfigure{
    \includegraphics[scale=0.3]{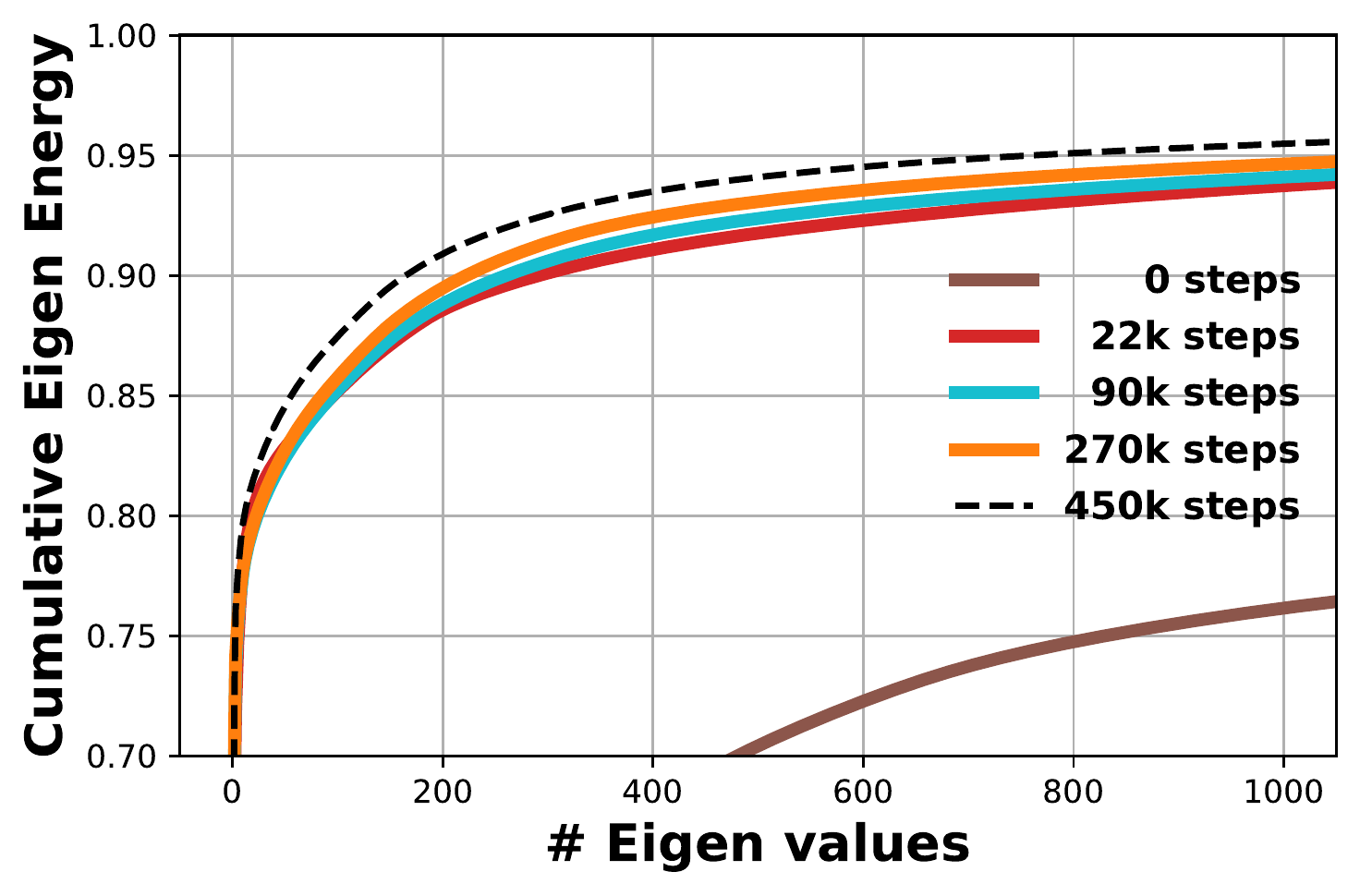}%
    }
    \subfigure{
    \includegraphics[scale=0.3]{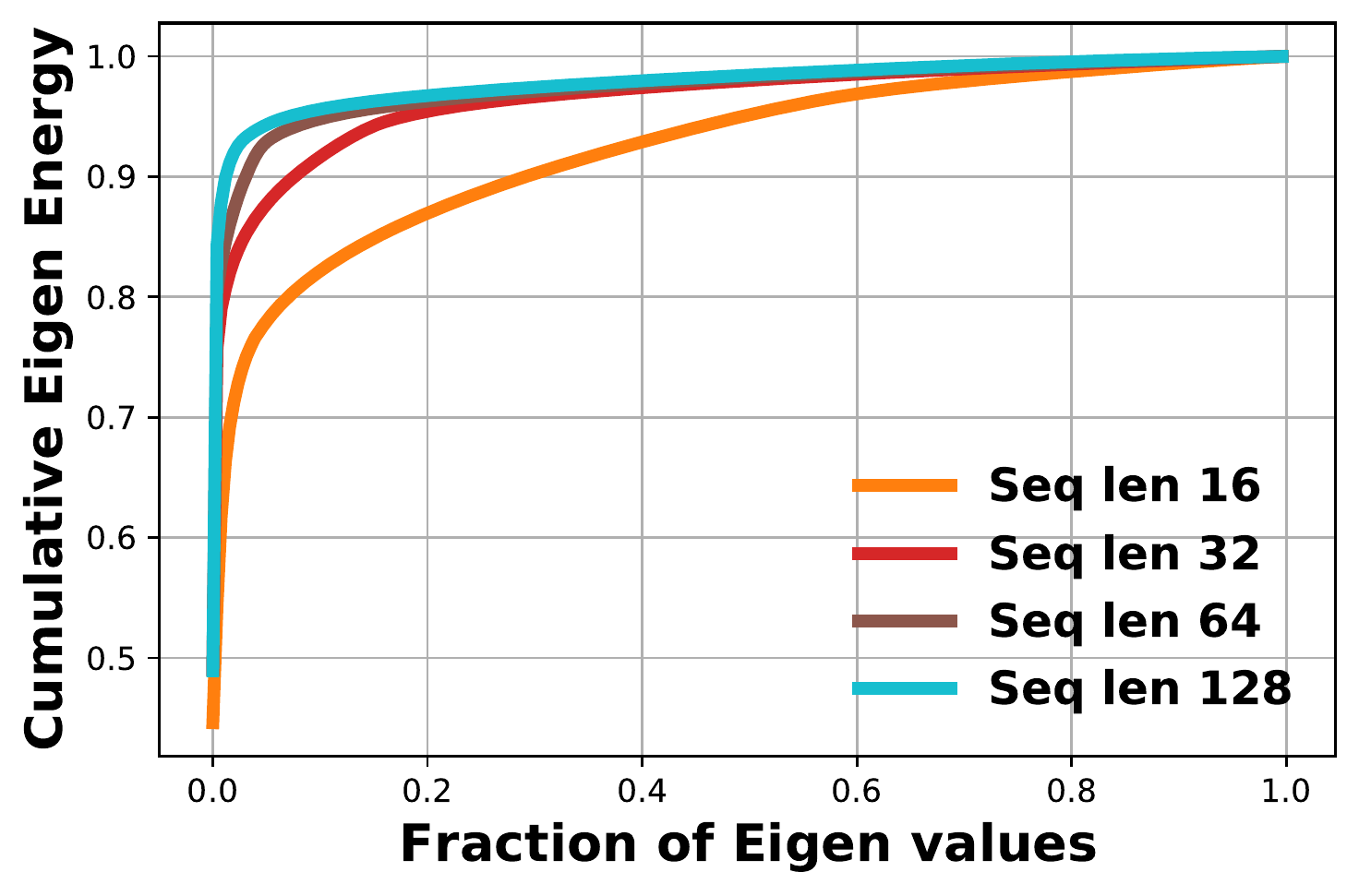}%
    } 
    \subfigure{
    \includegraphics[scale=0.3]{figs/bert_base_size_global_eigen.pdf}%
    } 
    \caption{\textbf{Eigen values of $\cov_a$}. Cumulative sum of eigen values of attention scores covariance matrix $\cov_a$ of a $\BB$ model  - Left: after varying number of training steps. We notice that the rank slightly decreases throughout training with a large reduction in the beginning. Middle: for different sequence length inputs. Note that x-axis here denotes the fraction of eigen values. Right: for varying model sizes. Note that rank slightly increases with model size.
    }
    \label{fig:eigen_bb_ablation}
    \vspace{-0.05in}
\end{figure*}
}


\newcommand{\insertFigBLEigenCross}{
\begin{figure*}[!t]
        \centering
    \subfigure{
    \includegraphics[scale=0.3]{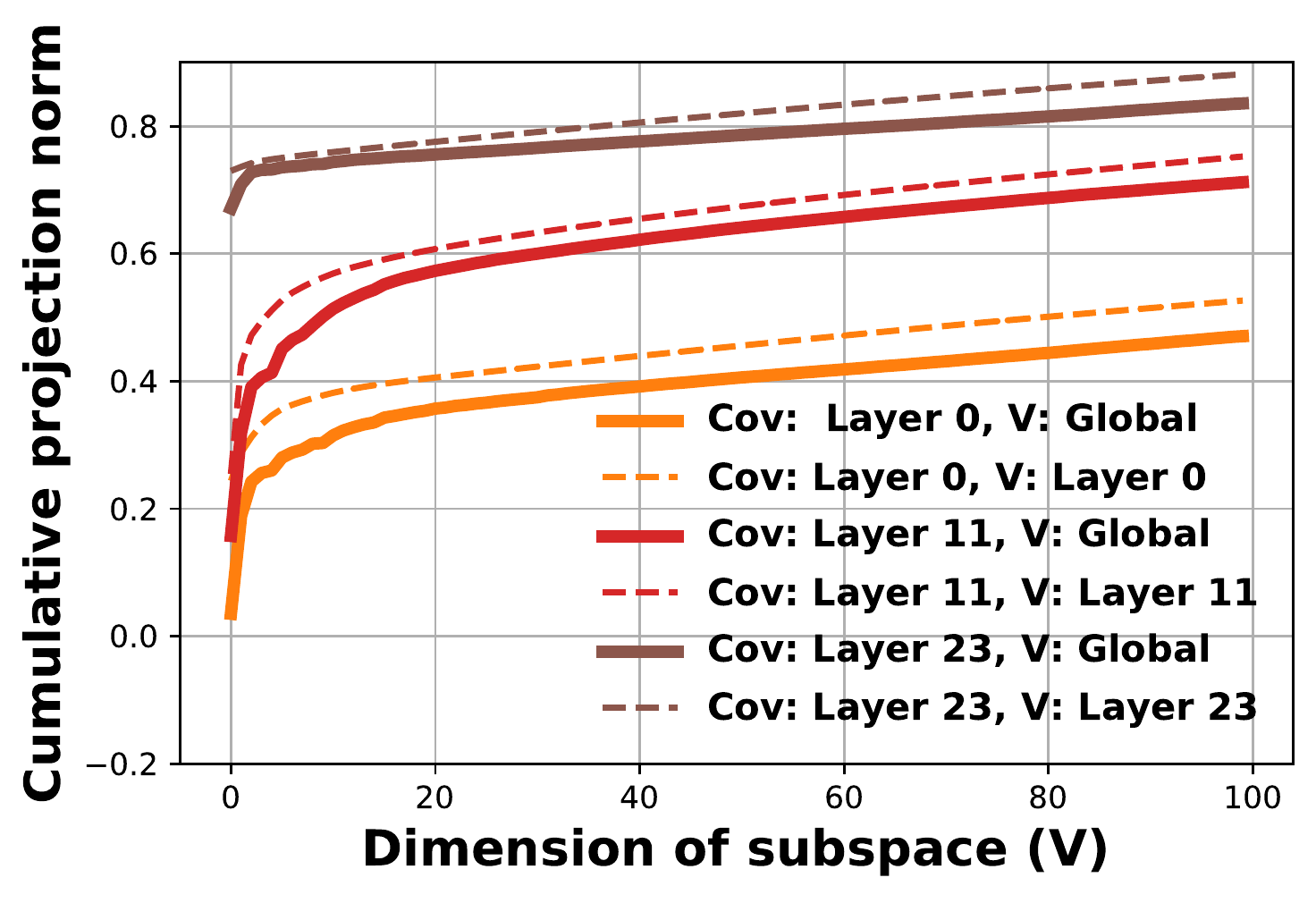}%
    } 
    \subfigure{
    \includegraphics[scale=0.3]{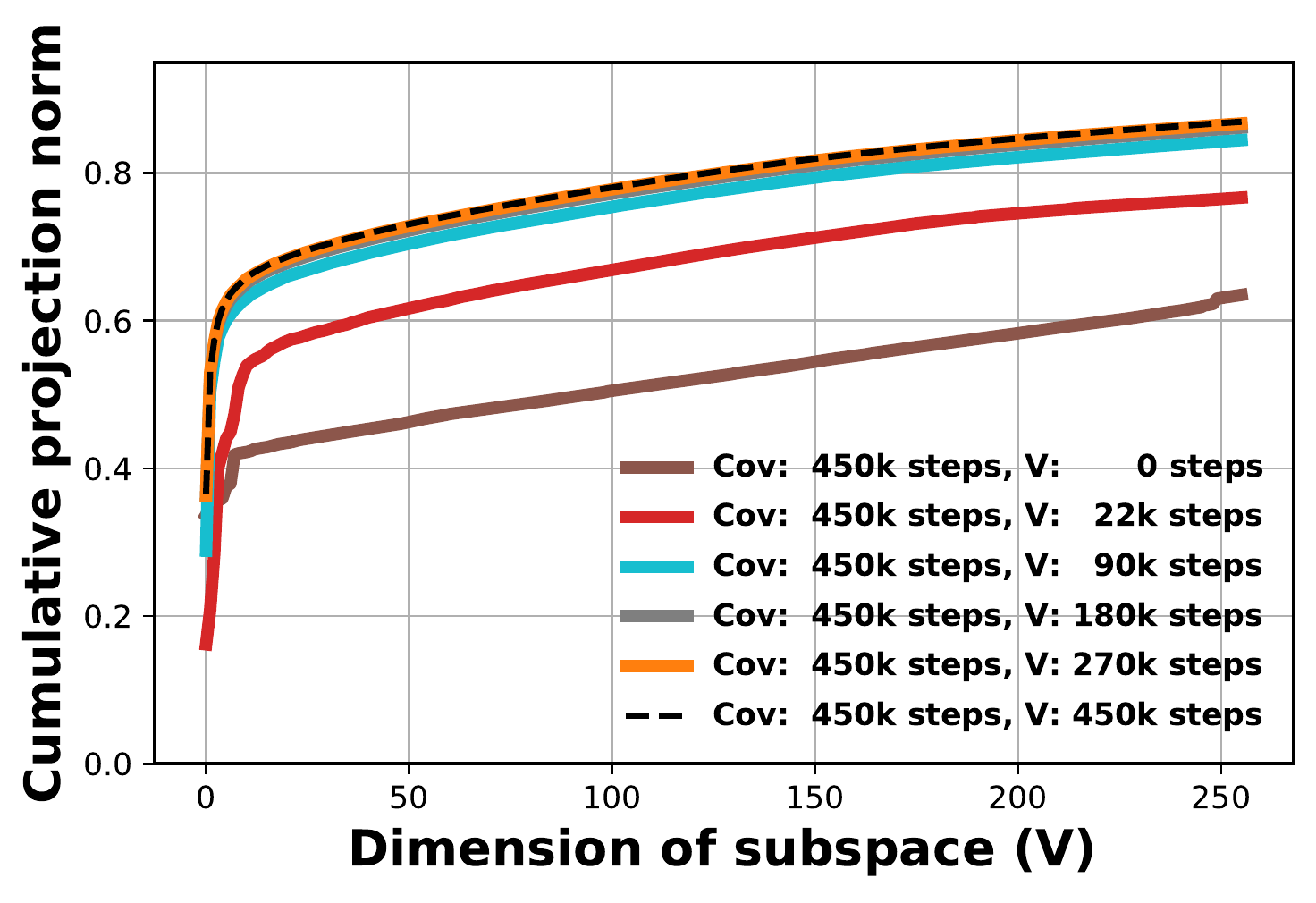}%
    }
    \caption{\textbf{Subspace similarity}. We plot the projection norm (cumulative energy eq~\ref{eq:projection}) to measure subspace overlap for different choices of attention scores $\cov_a$ and subspaces $\mV$. Left: Cumulative energy of covariance matrices ($\cov_a^l$) of different layers projected onto the top 256 eigen vectors of global covariance matrix $\cov_a$ of a $\BL$ model. We notice that there is substantial overlap in eigen subspaces of global and per layer attention scores. Right: Cumulative energy of covariance matrix after full training projected onto eigen vectors of covariance matrices after different numbers of training steps. We notice that the overlap increases quickly as training progresses. }
    \label{fig:eigen_bl_cross}
\end{figure*}
}

\newcommand{\insertFigBBEigenCross}{
\begin{figure*}[!t]
        \centering
    \subfigure{
    \includegraphics[scale=0.3]{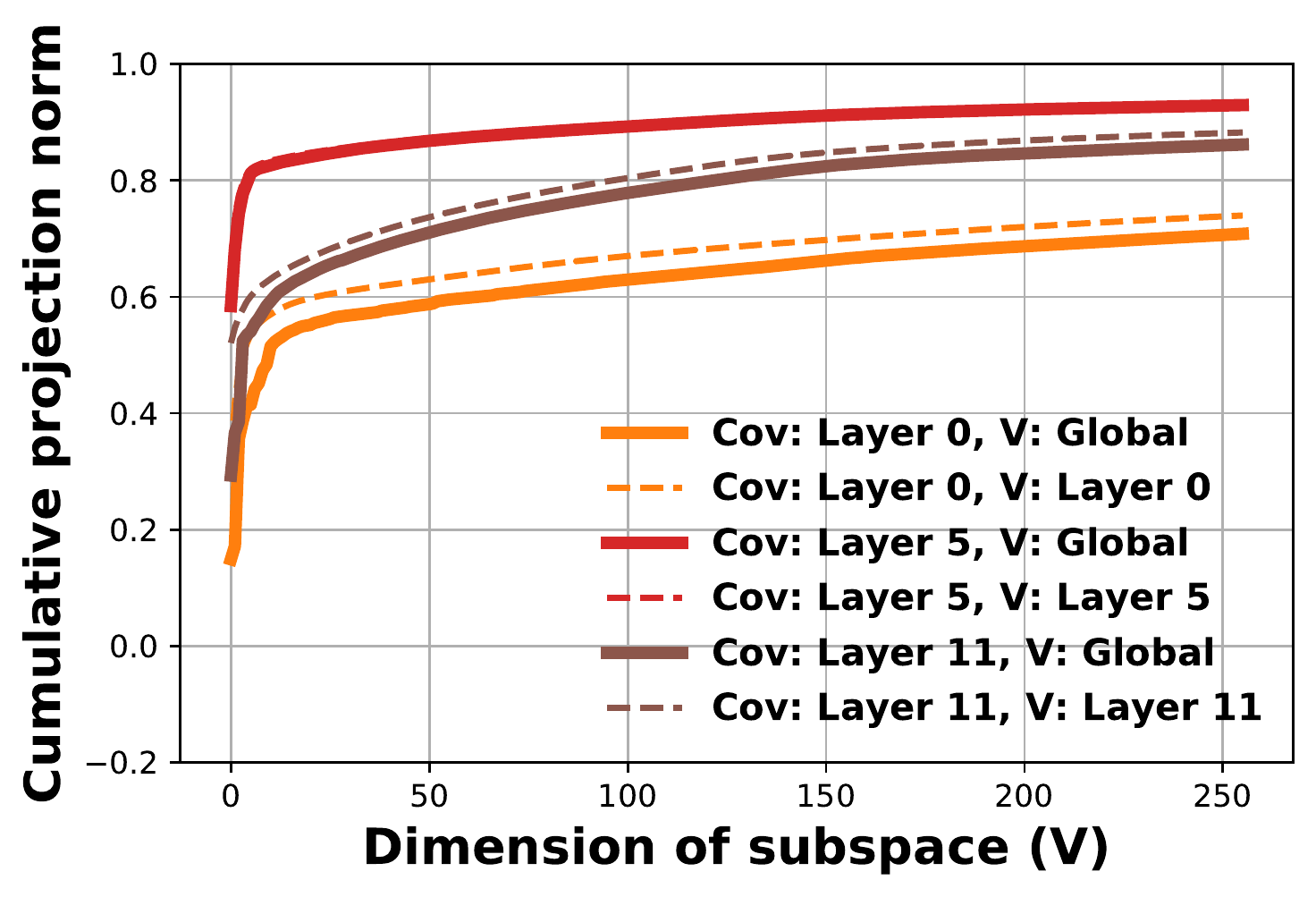}%
    } 
    \hfill
    \subfigure{
    \includegraphics[scale=0.3]{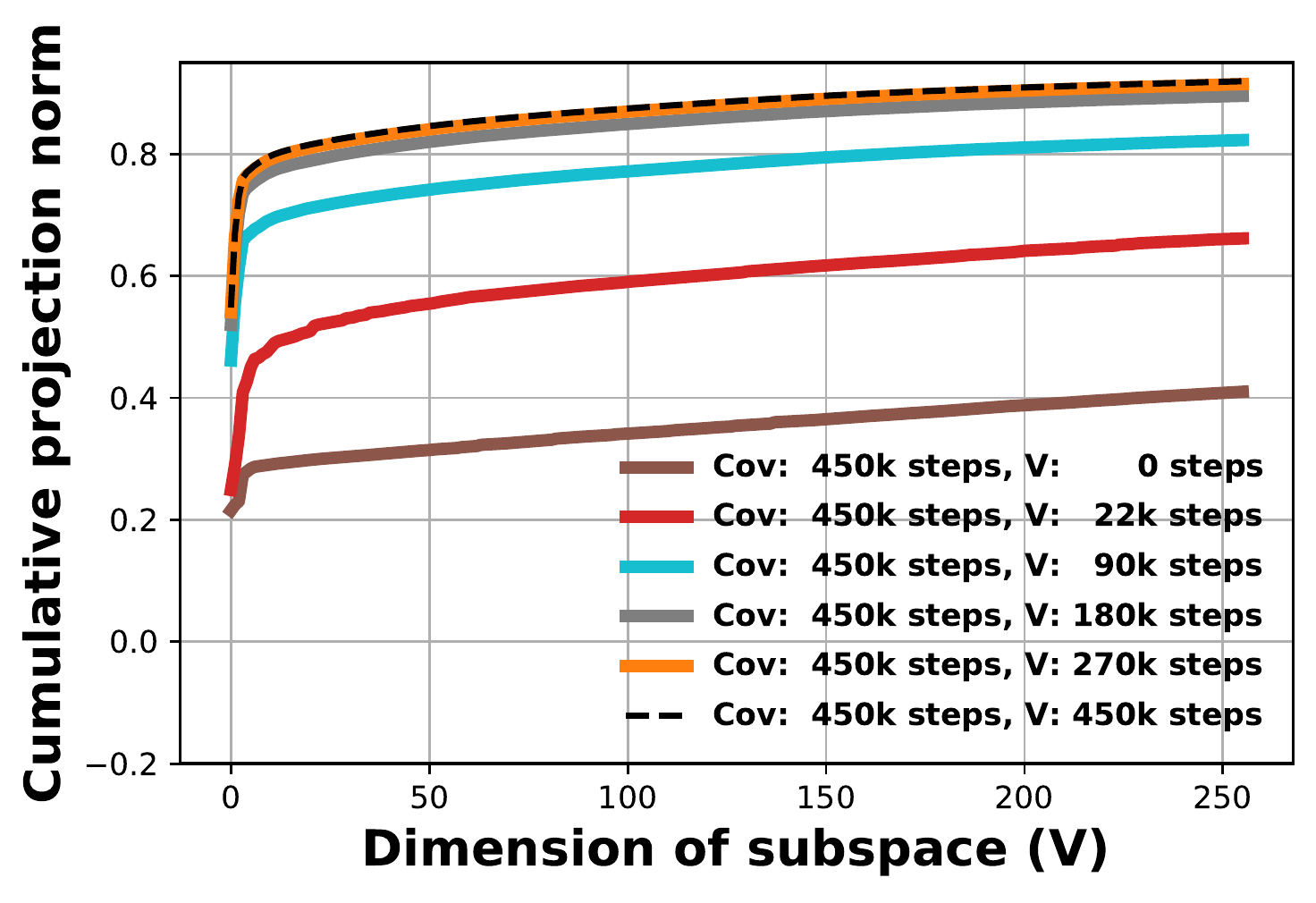}%
    }
    \hfill
    \subfigure{
    \includegraphics[scale=0.3]{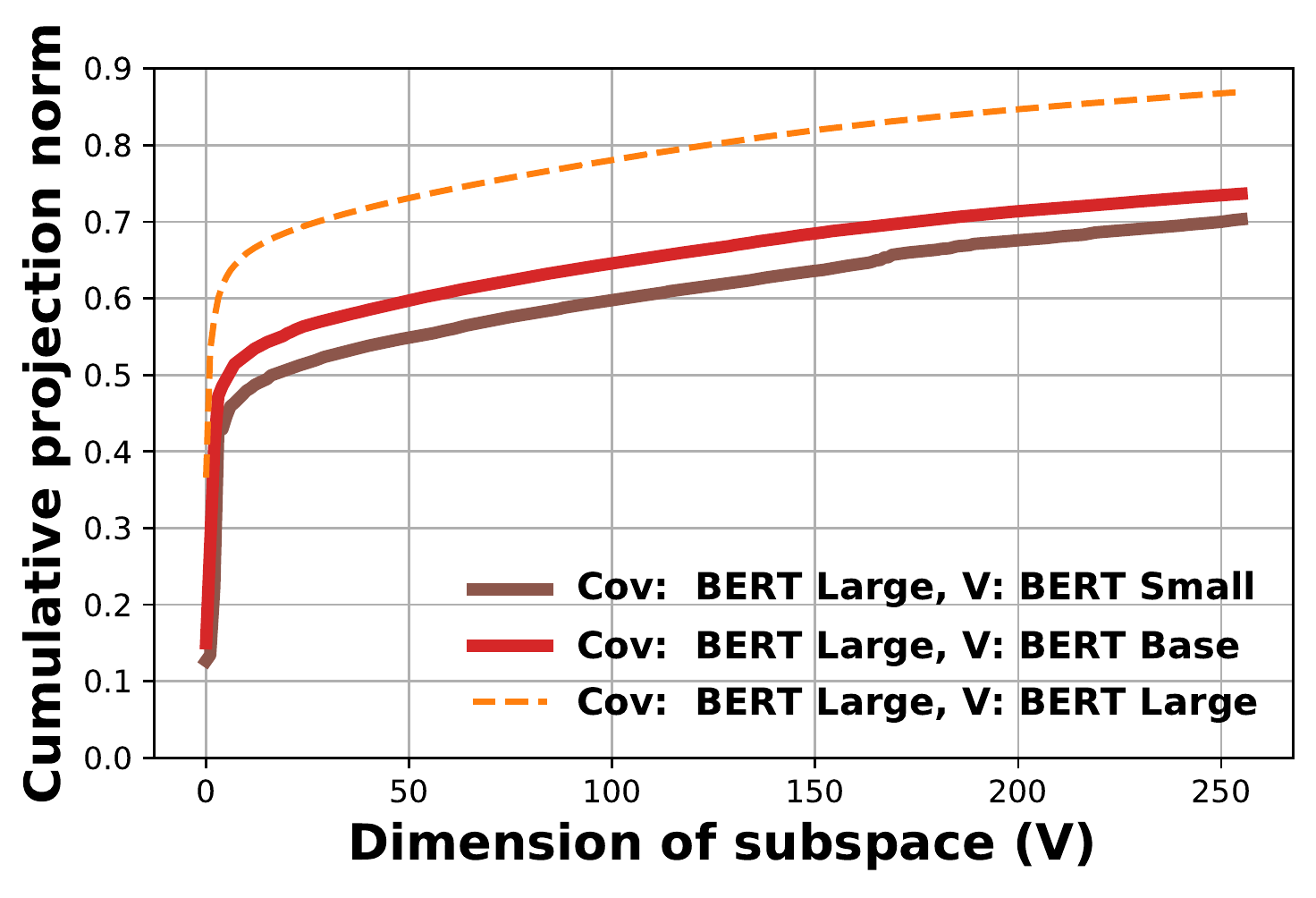}%
    } 
    \caption{\textbf{Subspace similarity}. We plot the cumulative projection norm (eq~\ref{eq:projection}) to measure subspace overlap for different choices of attention scores $\cov_a$ and subspaces $\mV$ for varying subspace dimension. Left: We project covariance matrices ($\cov_a^l$) of different layers projected onto the top 256 eigen vectors of global covariance matrix $\cov_a$ of a $\BB$ model. We notice that there is substantial overlap in eigen subspaces of global and per layer attention scores. Middle: We project attention scores covariance matrix after full training onto subspaces of attention scores after different number of training steps. We notice that the overlap increases quickly as training progresses. Right: Covariance matrix of a $\BL$ model projected onto eigen spaces of models with varying sizes.}
    \label{fig:eigen_bb_cross}
    \vspace{-0.05in}
\end{figure*}
}

\newcommand{\insertFigBBDatasets}{
\begin{figure*}[!t]
        \centering
    \subfigure{
    \includegraphics[scale=0.3]{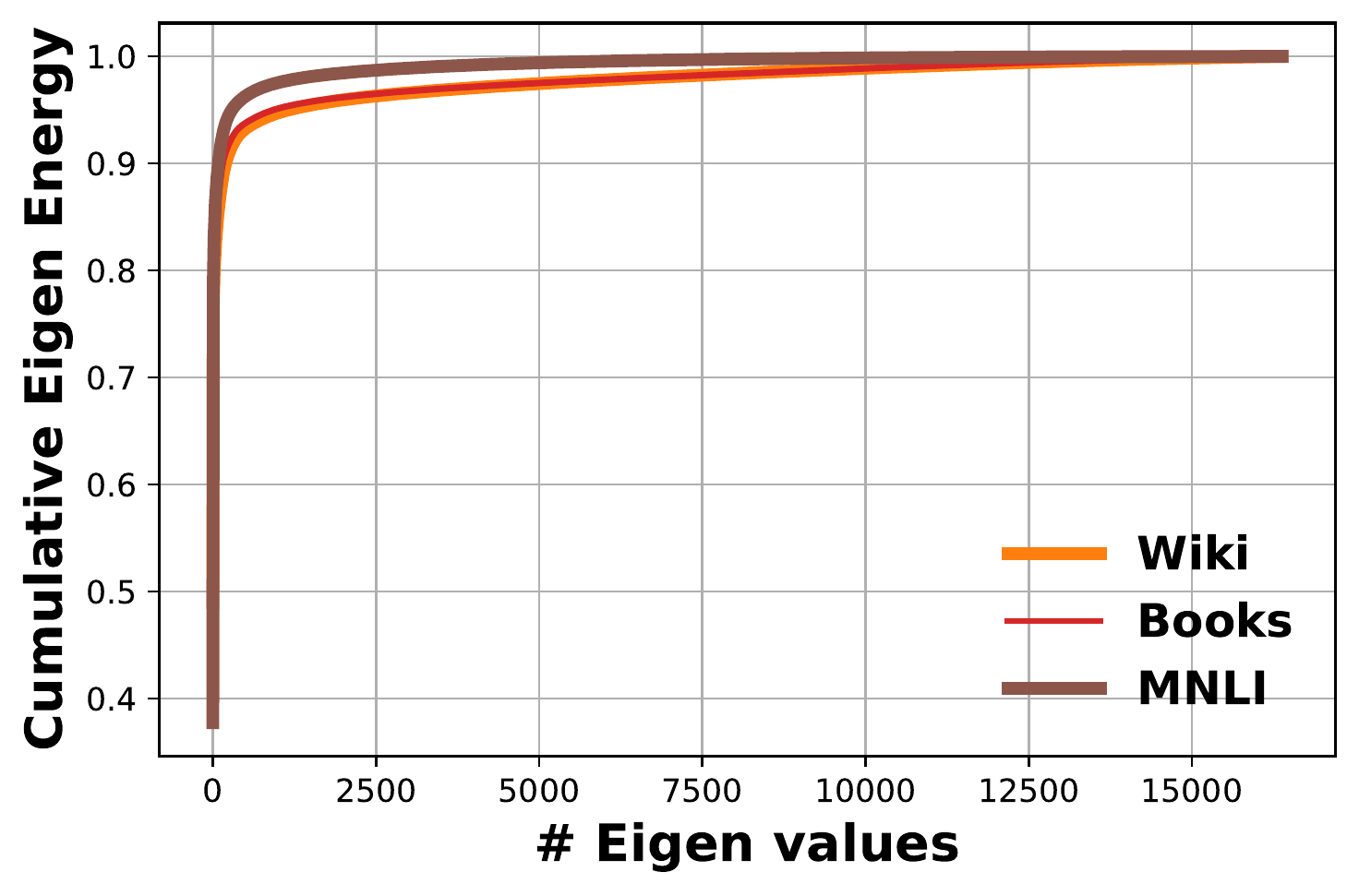}%
    } 
    \subfigure{
    \includegraphics[scale=0.3]{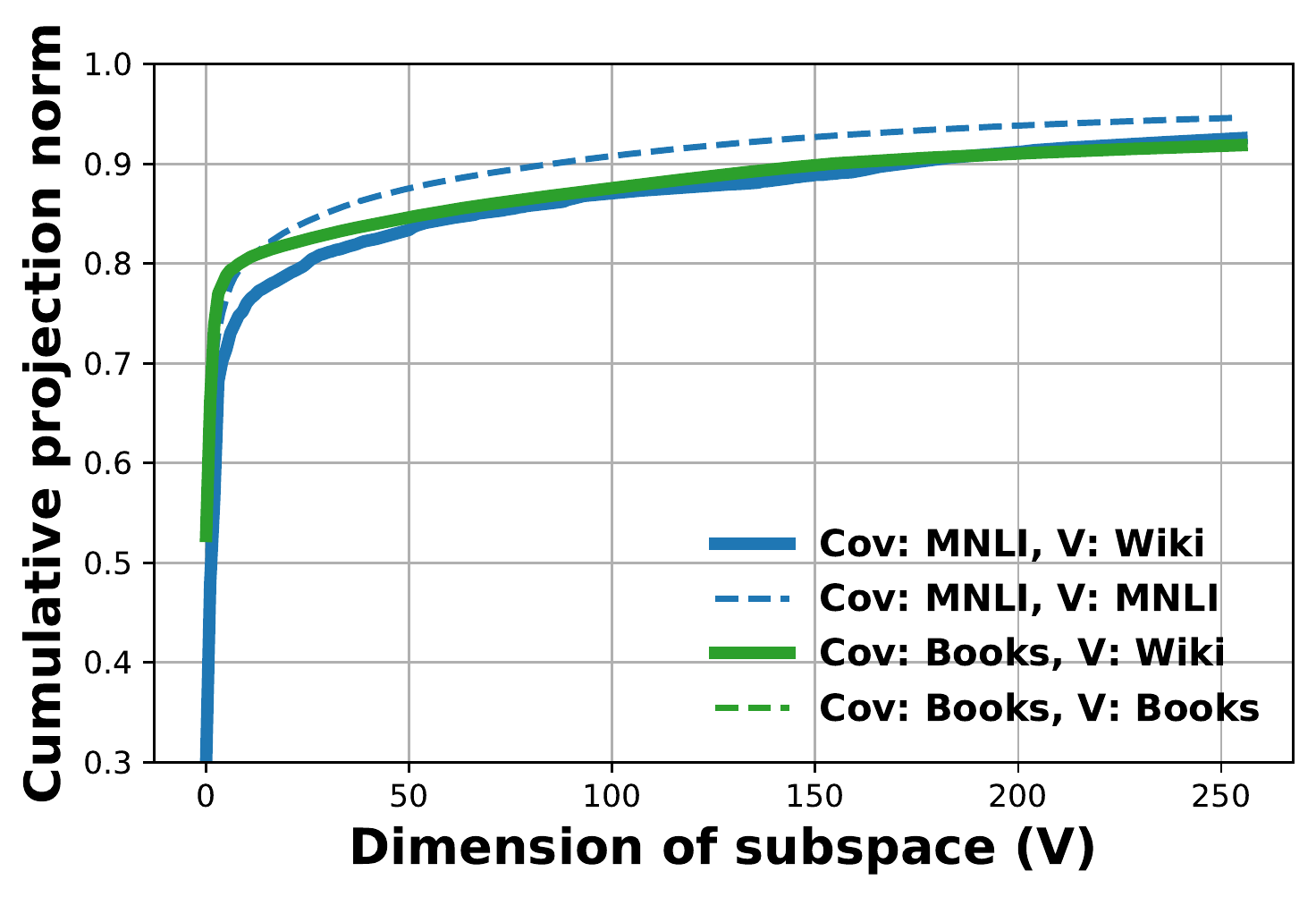}%
    } 
    \caption{\textbf{Variation across datasets}. Left: We plot the cumulative eigen spectrum of the attention scores variation on different datasets, and notice similar low dimensional structure.  Right: Similarity between subspaces (eq~\ref{eq:projection}) of attention scores from different datasets, for varying subspace dimension. We notice a large subspace overlap across different datasets.}
    \label{fig:eigen_bb_datasets}
    \vspace{-0.05in}
\end{figure*}
}

\newcommand{\insertFigBLGlobalEigenOneD}{
\begin{figure*}[!t]
    \subfigure{
    \includegraphics[scale=0.3]{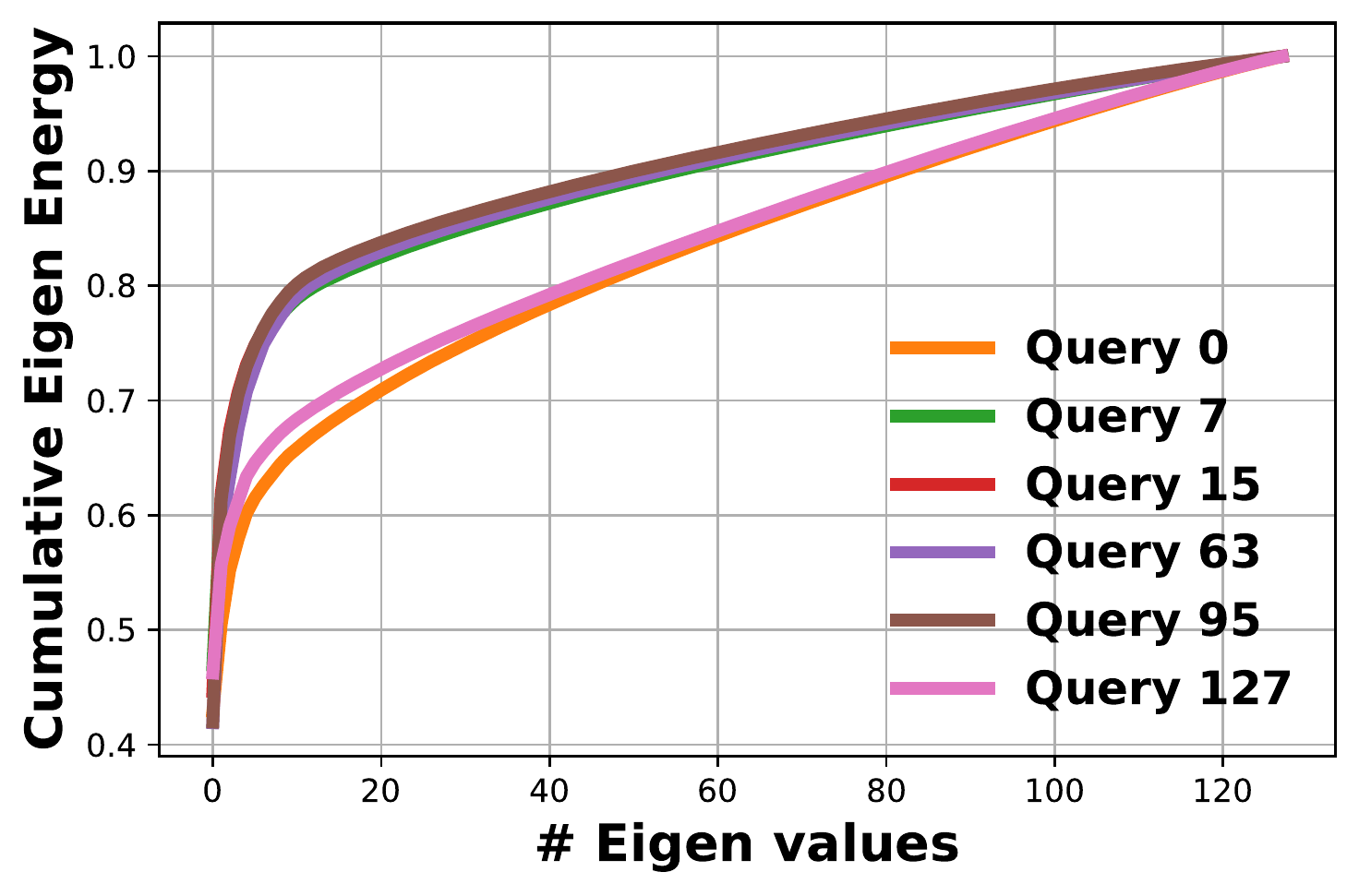}%
    }
    \hfill
    \subfigure{
    \includegraphics[scale=0.3]{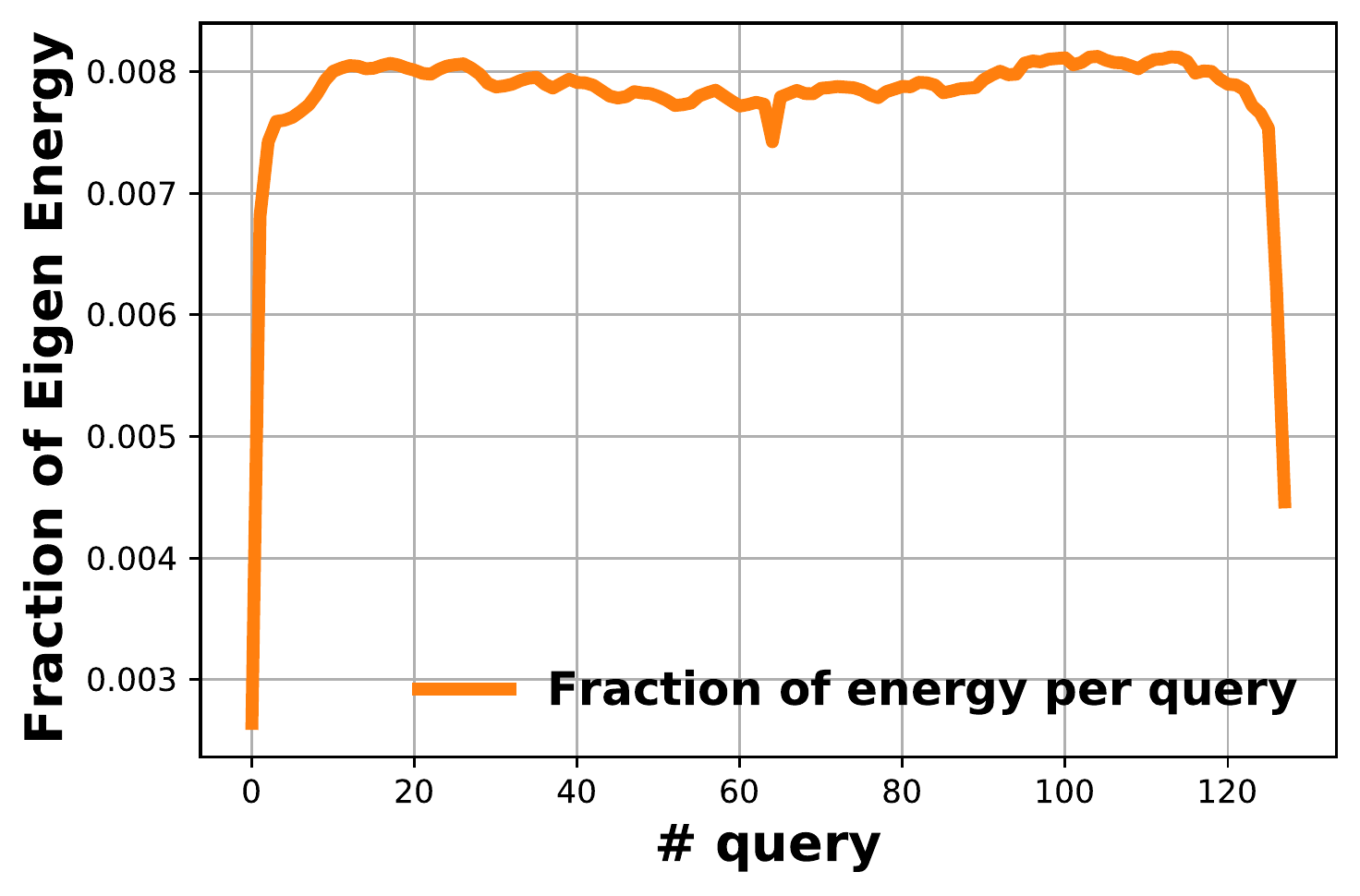}%
    }
    \hfill
    \subfigure{
    \includegraphics[scale=0.3]{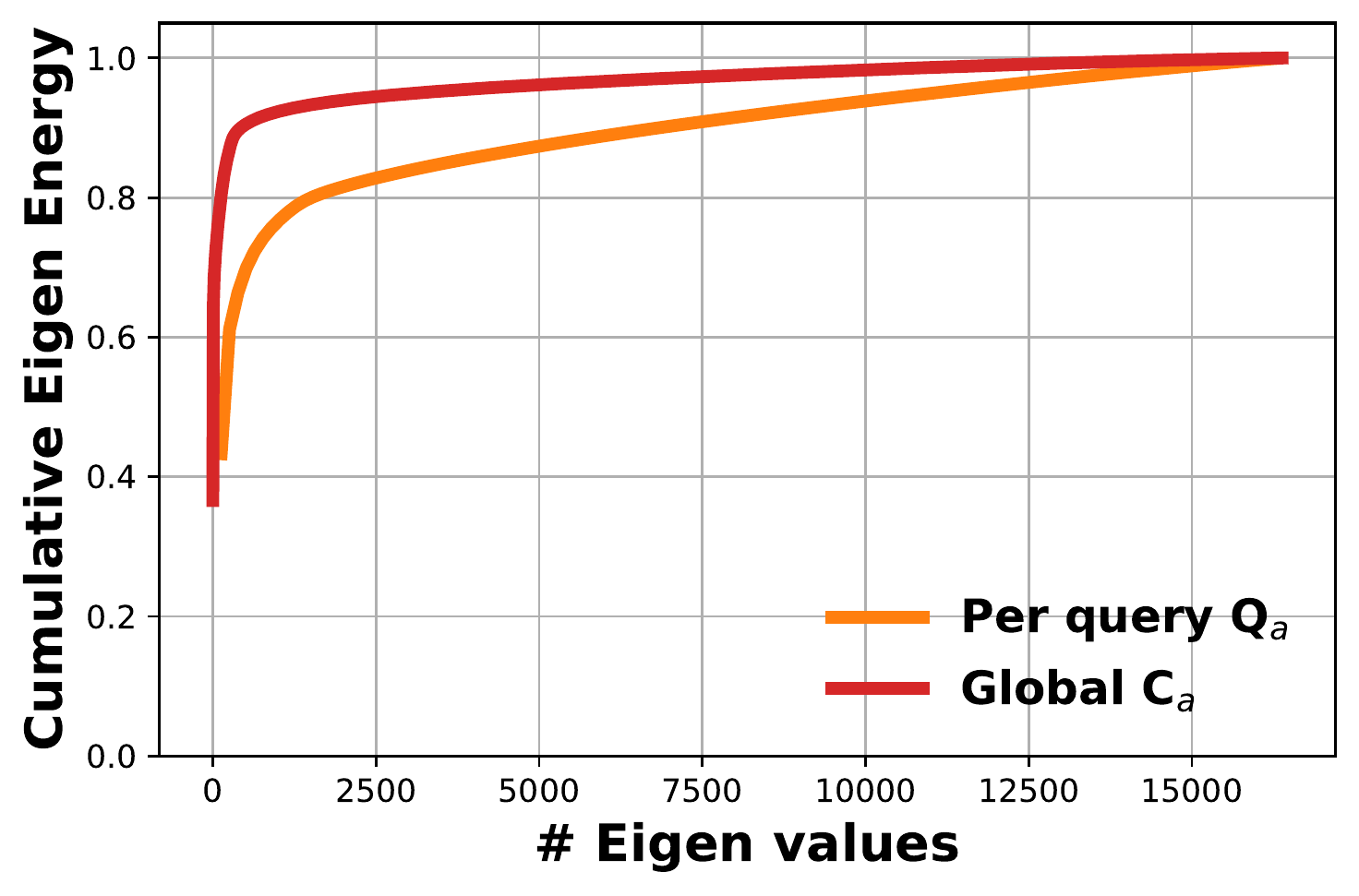}%
    }
    \caption{\textbf{Per-query eigen spectrum.} Left: Cumulative eigen values sum for different queries for a $\BL$ model. We notice that first and last queries, which are special tokens in BERT, have consistently higher rank compared to rest. Middle: Fraction of total eigenvalue sum for each query. We again notice that the special tokens have lower energy. Right: for each $k = i \times 128$, take top $i$ eigenvalues of each per-query covariance matrices. This is contrasted against the energy plot for the global attention scores patterns.}
    \label{fig:eigen_bl_global_1dim}
\end{figure*}
}

\newcommand{\insertFigBBGlobalEigenOneD}{
\begin{figure*}[!t]
    \subfigure{
    \includegraphics[scale=0.3]{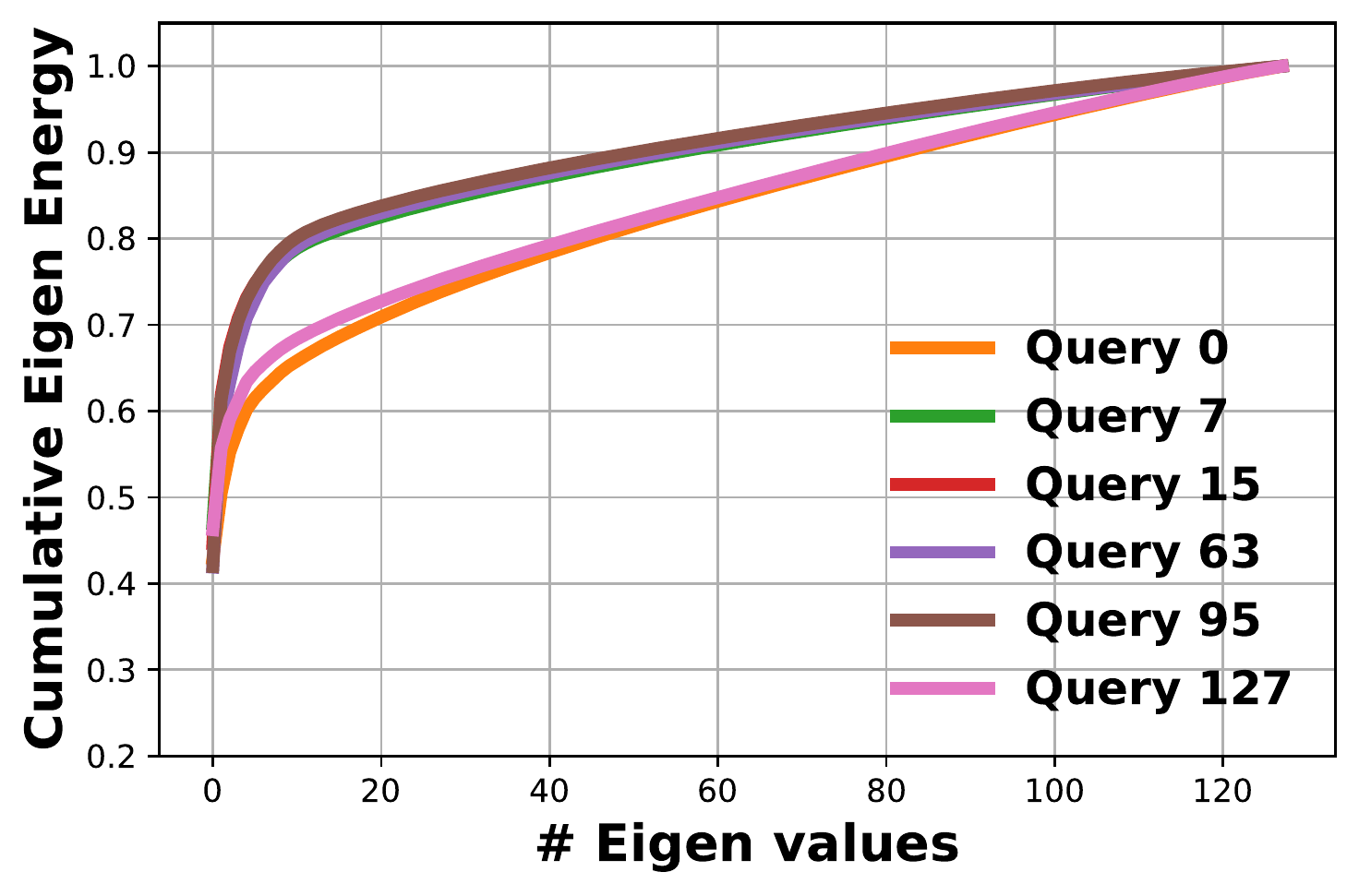}%
    }
    \hfill
    \subfigure{
    \includegraphics[scale=0.3]{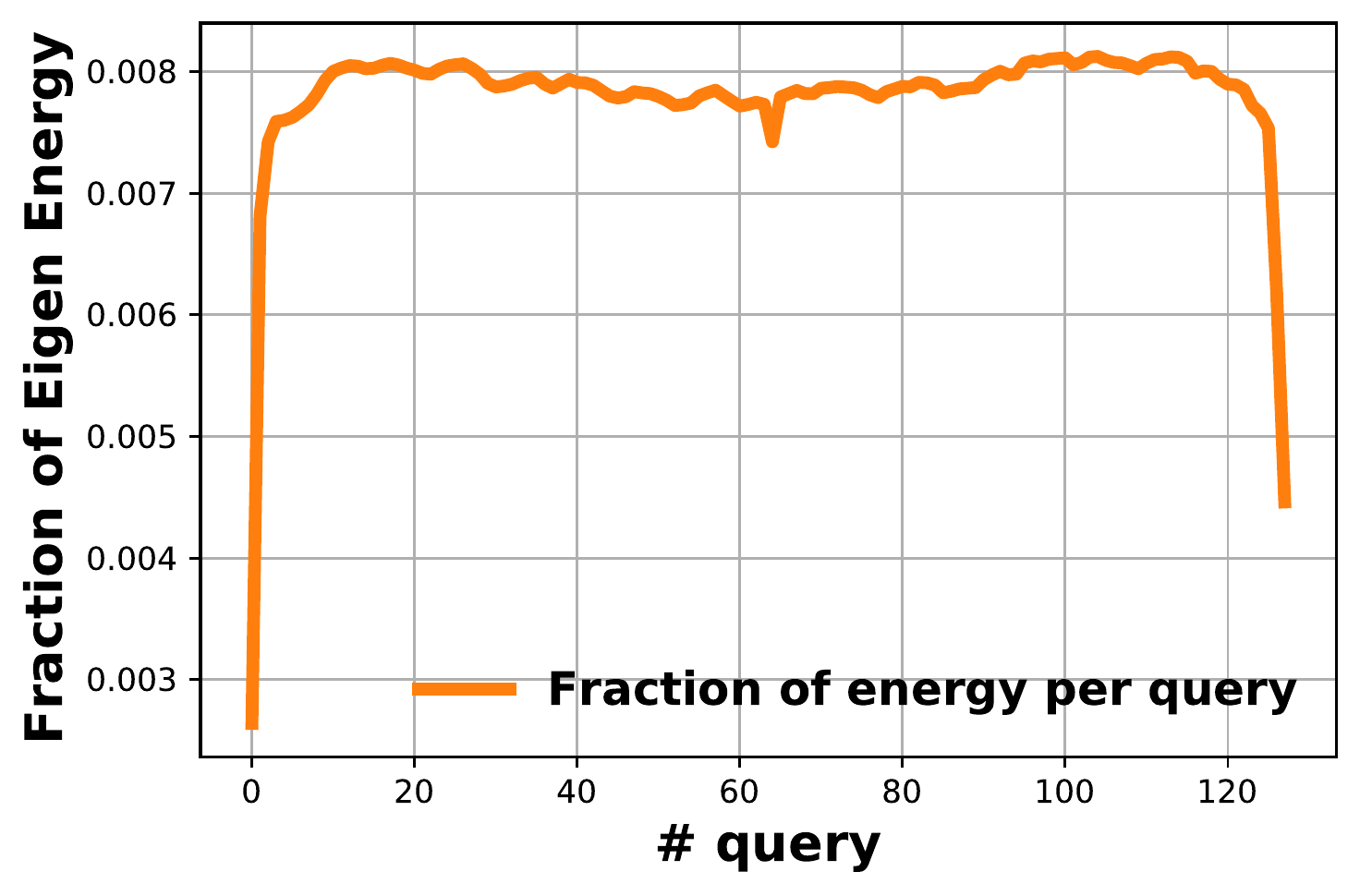}%
    }
    \hfill
    \subfigure{
    \includegraphics[scale=0.3]{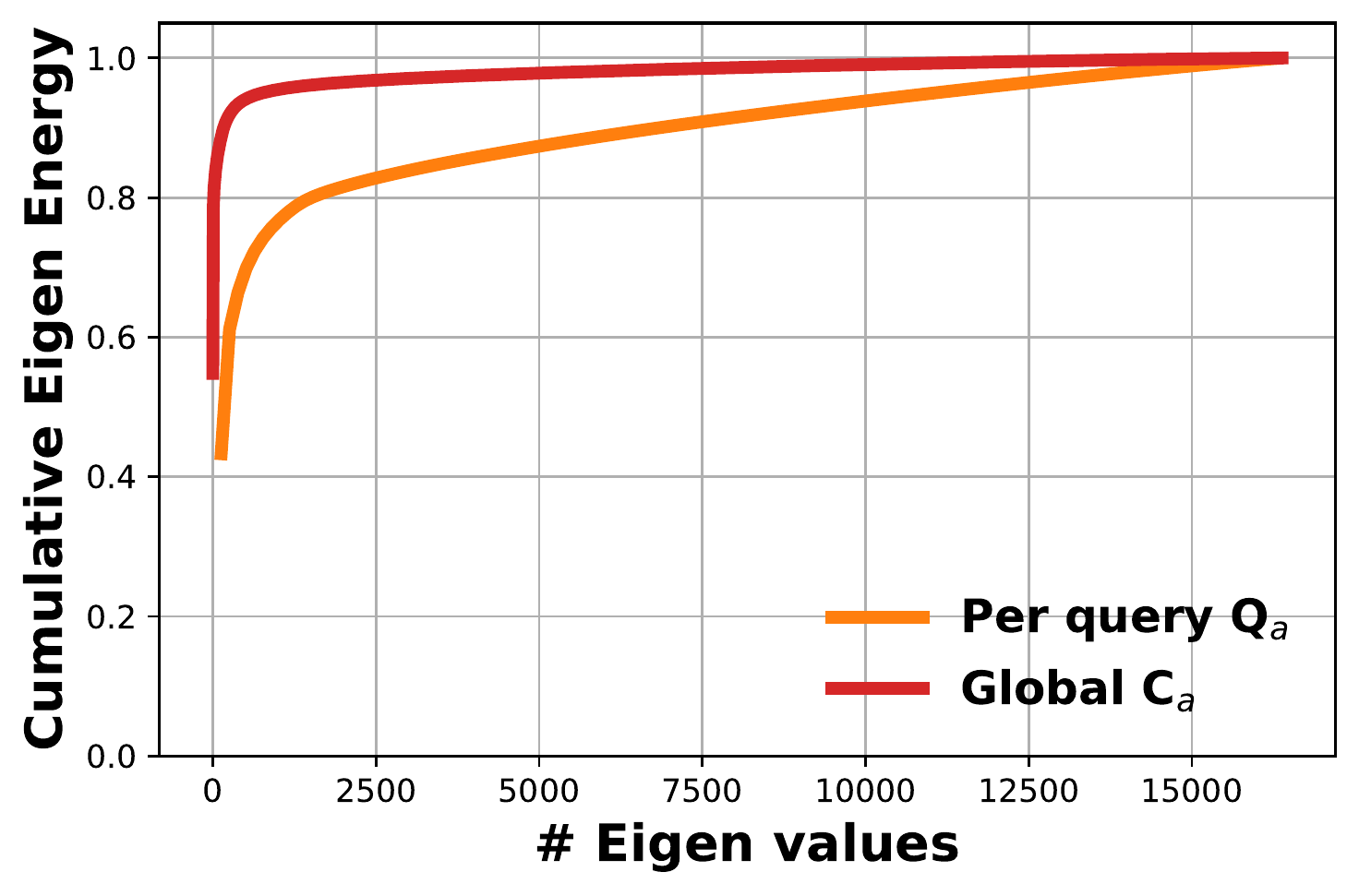}%
    }
    \caption{\textbf{Per-query eigen spectrum.} Left: Cumulative eigen values sum for different queries. We notice that first and last queries, which are special tokens in BERT, have consistently higher rank compared to rest. Middle: Fraction of total eigenvalue sum for each query. We again notice that the special tokens have lower energy. Right: for each $k = i \times 128$, we take top $i$ eigenvalues of each per-query covariance matrix ($\qcov_a$), and plot their cumulative eigen spectrum. This is contrasted against the eigen spectrum of the global attention scores ($\cov_a$). We notice that per-query attention scores capture majority of the variation in the global attention scores.}
    \vspace{-0.05in}
    \label{fig:eigen_bb_global_1dim}
\end{figure*}
}

\newcommand{\insertFigBLGlobalEigenOneDVisualize}{
\begin{figure*}[!t]
    \centering
    \subfigure{
    \includegraphics[scale=0.3]{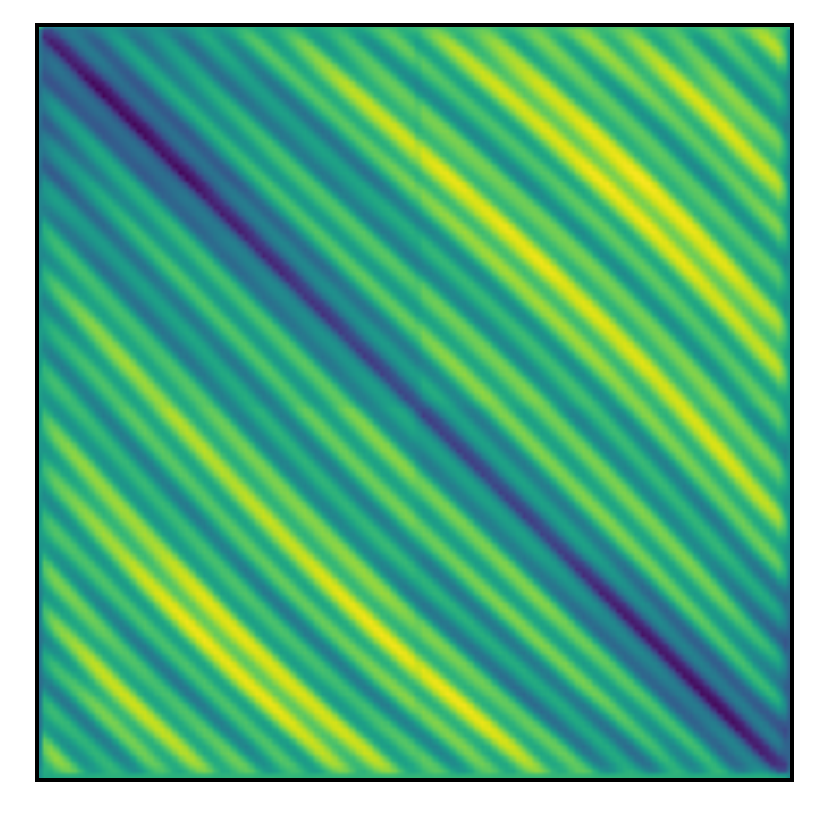}%
    } 
    \hfill
    \subfigure{
    \includegraphics[scale=0.3]{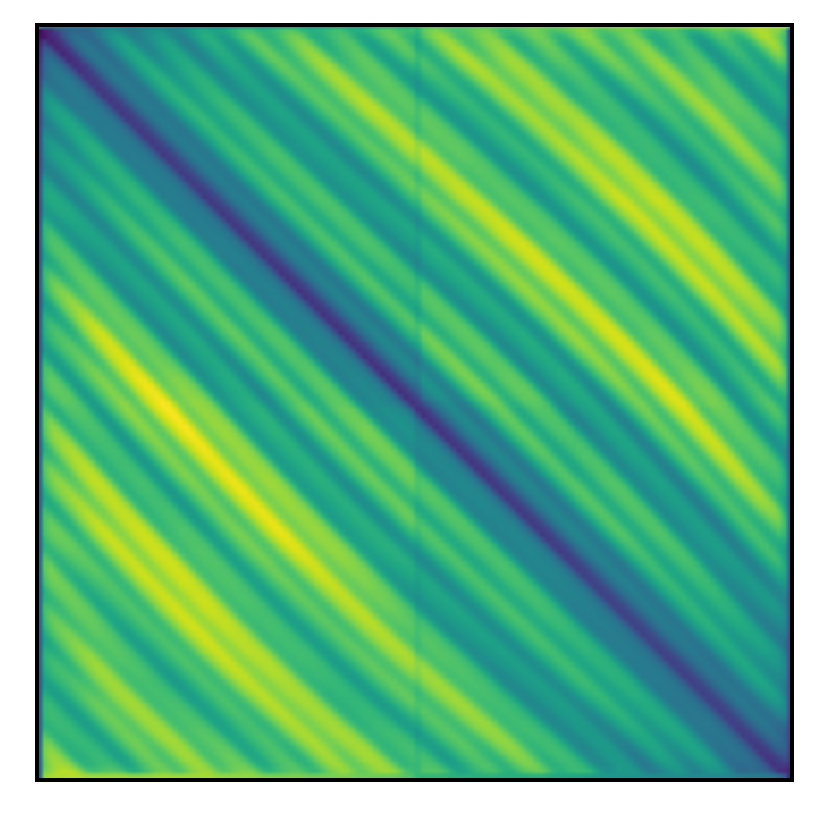}%
    }
    \hfill
    \subfigure{
    \includegraphics[scale=0.3]{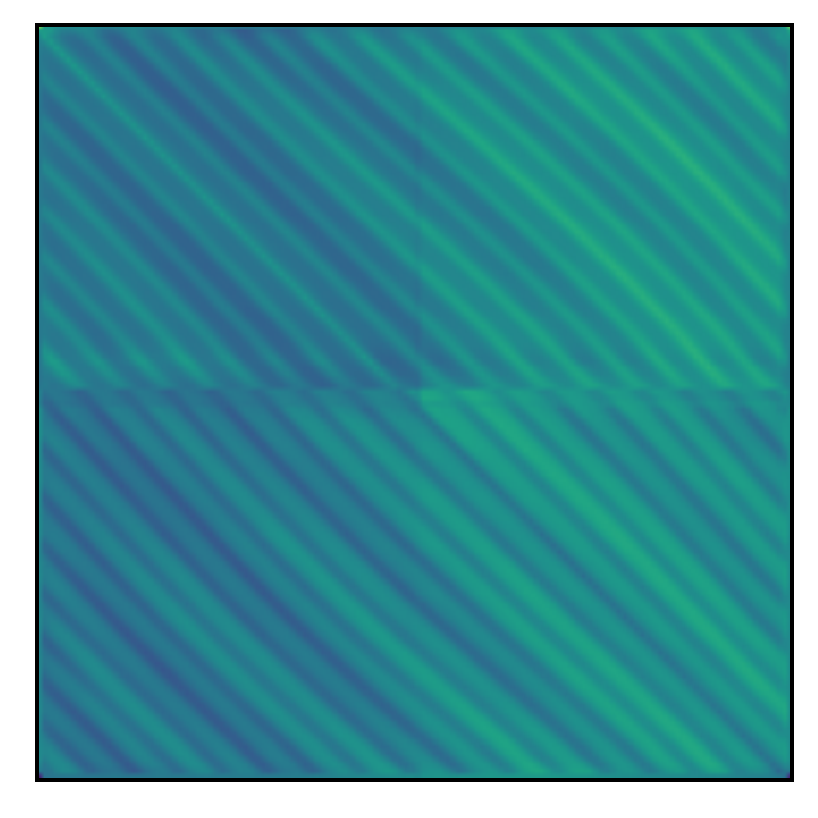}%
    }
    \hfill
    \subfigure{
    \includegraphics[scale=0.3]{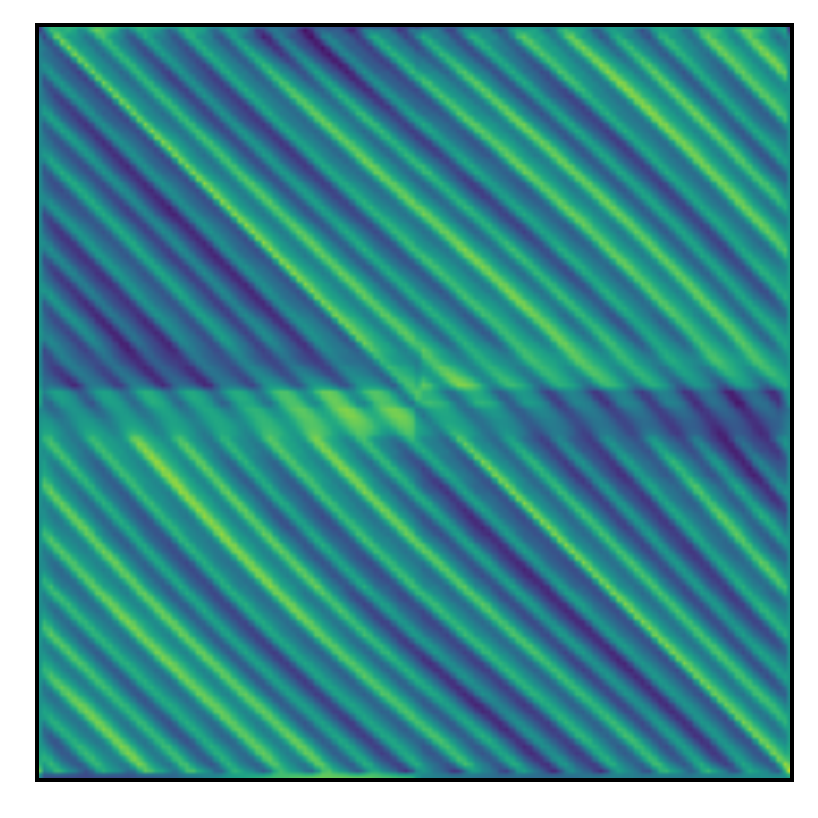}%
    }
    \hfill
    \subfigure{
    \includegraphics[scale=0.3]{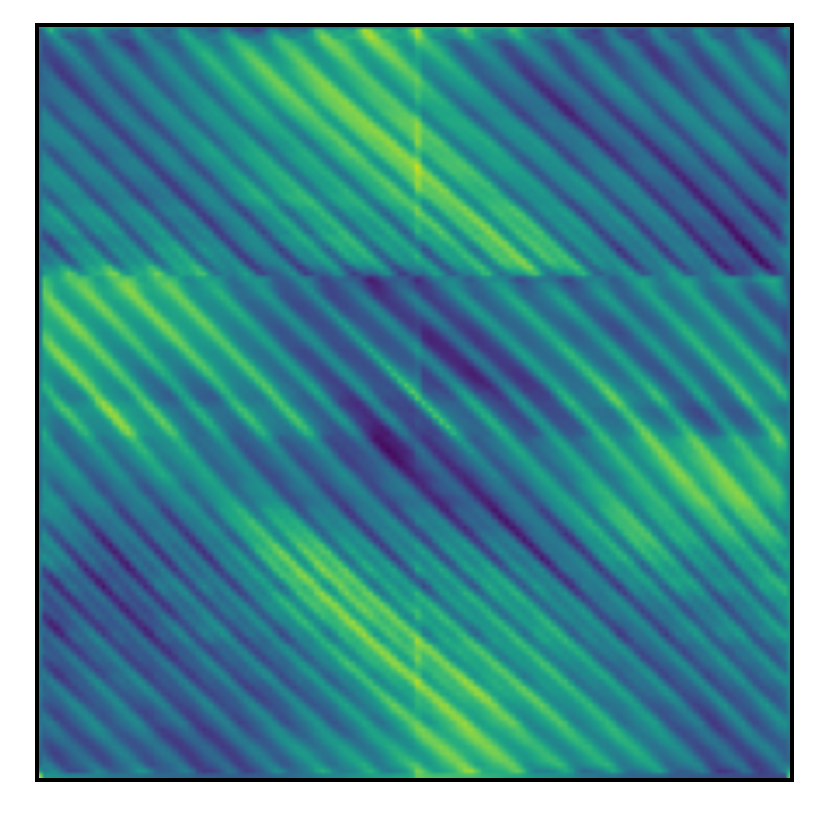}%
    }
    \caption{\textbf{Per-query principal components}. Visualization of per-query attention score patterns from the top 5 eigenvectors (1st for the leftmost Figure, then 2nd etc.) from each query stacked into their corresponding rows for a $\BL$ model. Note the similarities between the top eigenvectors of the per-query attention scores and global attention scores (Fig~\ref{fig:eigen_bl_global_2dim_visualize}).}
    \label{fig:eigen_bl_global_1dim_visualize}
\end{figure*}
}

\newcommand{\insertFigBBGlobalEigenOneDVisualize}{
\begin{figure*}[!t]
    \centering
    \subfigure{
    \includegraphics[scale=0.3]{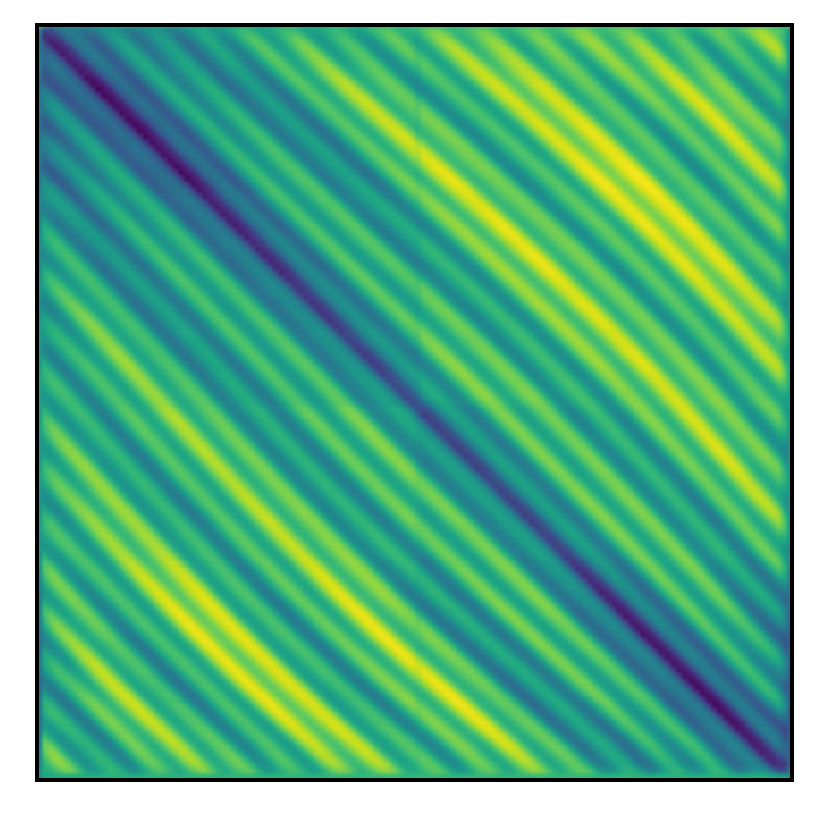}%
    } 
    \hfill
    \subfigure{
    \includegraphics[scale=0.3]{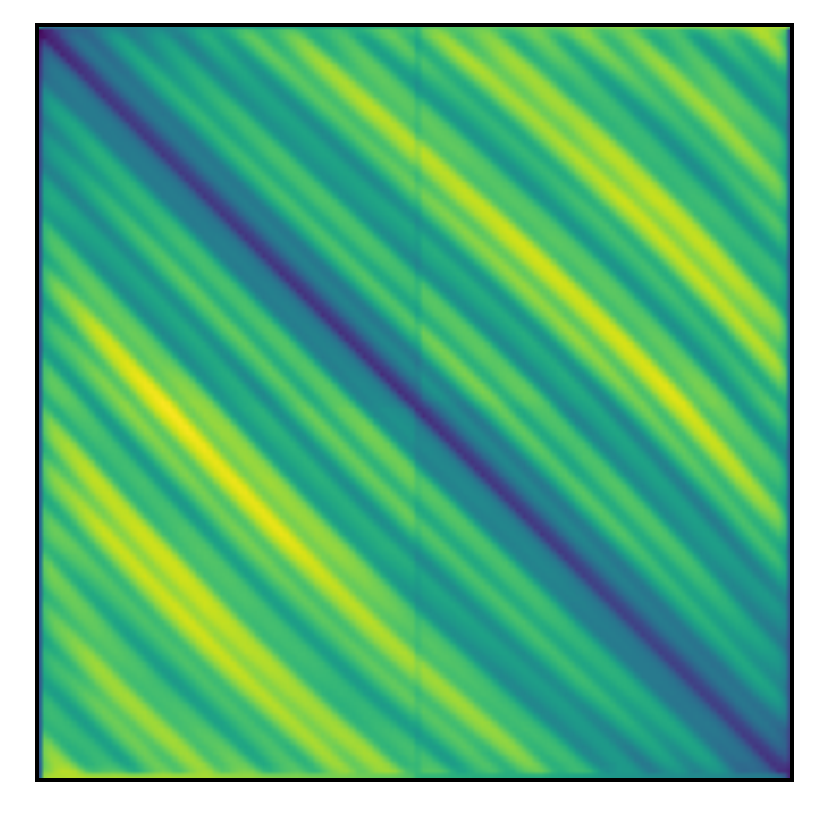}%
    }
    \hfill
    \subfigure{
    \includegraphics[scale=0.3]{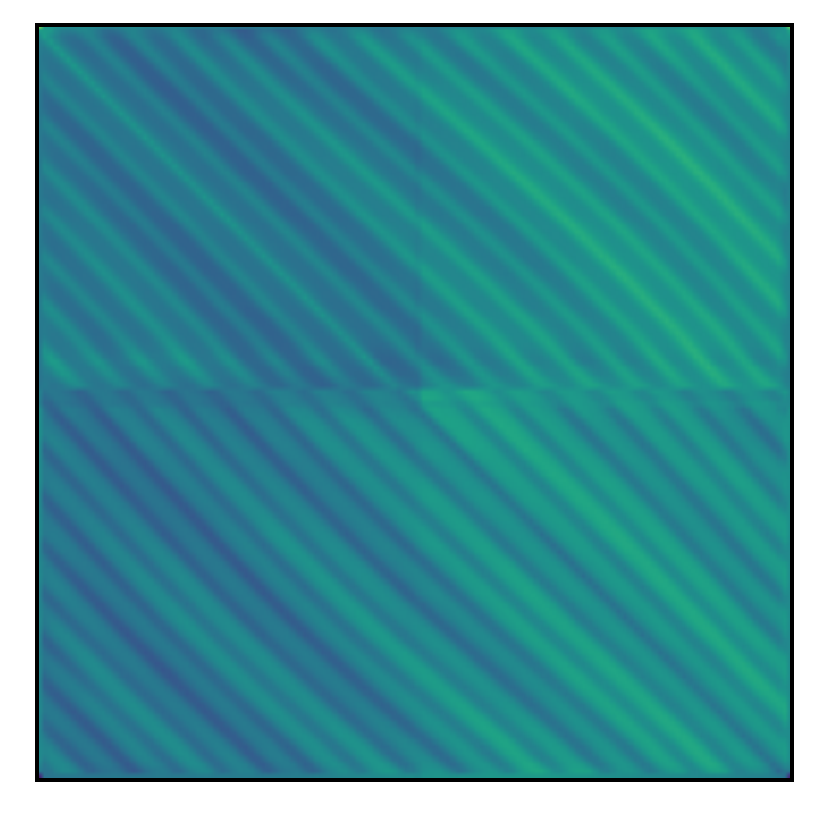}%
    }
    \hfill
    \subfigure{
    \includegraphics[scale=0.3]{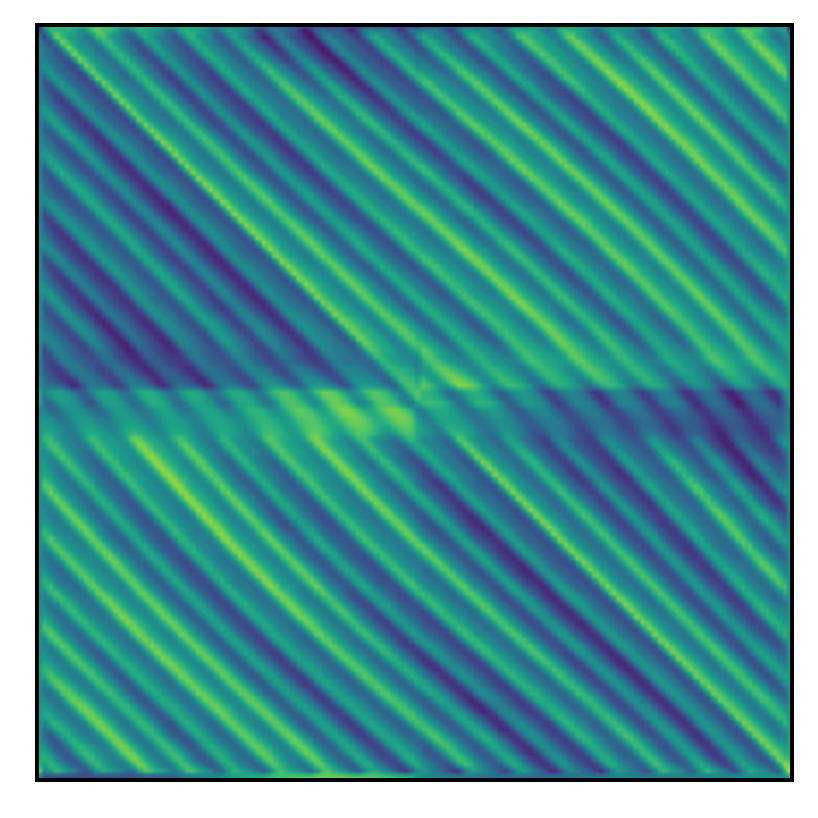}%
    }
    \hfill
    \subfigure{
    \includegraphics[scale=0.3]{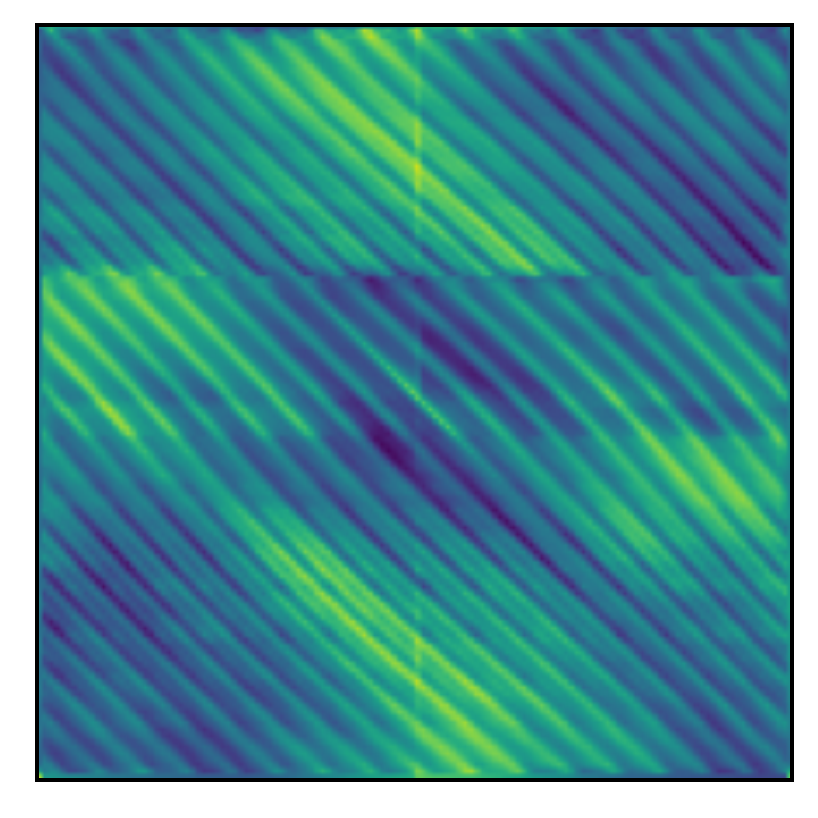}%
    }
    \caption{\textbf{Per-query principal components}. Visualization of per-query attention score patterns from the top 5 eigenvectors (1st for the leftmost Figure, then 2nd etc.) from each query stacked into their corresponding rows. Note the similarities between the top eigenvectors of the per-query attention scores and global attention scores (Fig~\ref{fig:eigen_bb_global_2dim_visualize}).}
    \label{fig:eigen_bb_global_1dim_visualize}
    \vspace{-0.1in}
\end{figure*}
}

\newcommand{\insertFigBBGlobalEigenTwoDVisualize}{
\begin{figure*}[!t]
    \centering
    \includegraphics[width=\textwidth]{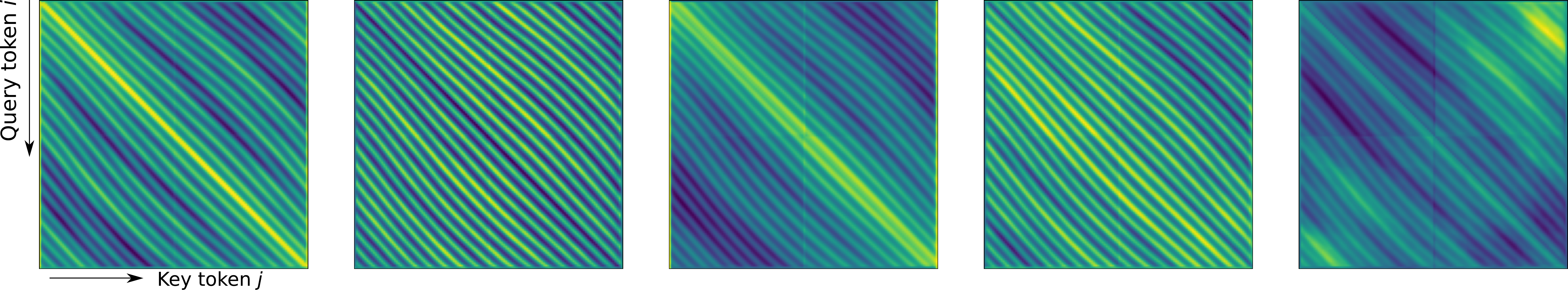}%
    \caption{\textbf{Global principal components}. Visualization of attention score patterns captured by the top 5 eigenvectors of $\cov_a$ of a $\BB$ model. We notice that the leading principal components capture predominantly shifted diagonal patterns.}
    \label{fig:eigen_bb_global_2dim_visualize}
\end{figure*}
}

\newcommand{\insertFigBLGlobalEigenTwoDVisualize}{
\begin{figure*}[!t]
    \centering
    \includegraphics[width=\textwidth]{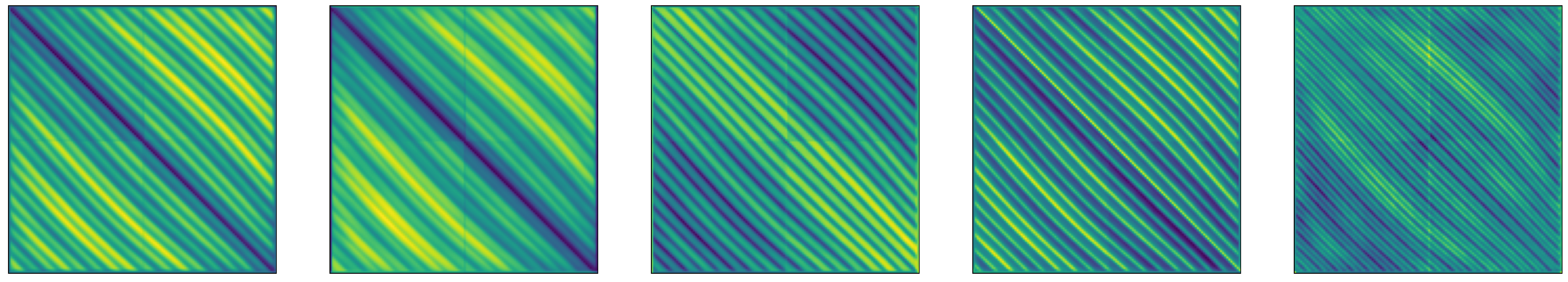}%
    \caption{\textbf{Global principal components}. Visualization of attention score patterns captured by the top 5 eigenvectors of $\cov_a$ of a $\BL$ model. We notice that the leading principal components capture predominantly shifted diagonal patterns.}
    \label{fig:eigen_bl_global_2dim_visualize}
\end{figure*}
}

\begin{abstract}
State-of-the-art transformer models use pairwise dot-product based self-attention, which comes at a computational cost quadratic in the input sequence length. In this paper, we investigate the global structure of attention scores computed using this dot product mechanism on a typical distribution of inputs, and study the principal components of their variation. Through eigen analysis of full attention score matrices, as well as of their individual rows, we find that most of the variation among attention scores lie in a low-dimensional eigenspace. Moreover, we find significant overlap between these eigenspaces for different layers and even different transformer models. Based on this, we propose to compute scores only for a partial subset of token pairs, and use them to estimate scores for the remaining pairs. Beyond investigating the accuracy of reconstructing attention scores themselves, we investigate training transformer models that employ these approximations, and analyze the effect on overall accuracy. Our analysis and the proposed method provide insights into how to balance the benefits of exact pair-wise attention and its significant computational expense.

\end{abstract}

\section{Introduction}

Transformers deliver state-of-the-art performance in several tasks in natural language processing~\citep{vaswani2017attention, devlin2018bert, radford2018gpt, radford2019gpt2,raffel2020exploring}, and are beginning to show promise in other domains~\cite{dosovitskiy2020image}. Their success is attributed to their self-attention mechanism. While previous architectures based on convolutional or recurrent layers had a-priori fixed structures for interaction between different positions in the input, self-attention in transformers allows arbitrary input-dependent interaction through attention scores based on dot product similarities between learned hidden feature representations. The importance of this flexible input-dependent attention has been established both empirically~\citep{wu2019pay, raganato2020fixed, tay2020synthesizer} and theoretically~\cite{yun2019transformers}.

While transformers have the expressive capacity to output arbitrary attention scores between all the input token pairs, these scores are computed based on the inputs to these models. In the case of natural language, we know that these inputs are structured and have a more restricted distribution than random sequences of words~\cite{brown1992estimate}. It follows, then, that attention scores computed from inputs drawn from a structured natural distribution will themselves exhibit some structure. In this paper, we analyze the distribution of attention scores computed by transformer models from natural language. 

Our analysis shows that the attention scores---either the global attention matrix of all scores across all pairs of tokens, or scores corresponding to a given query token position---are not arbitrary and much of their variability can be explained by a relatively small number of principal components. Moreover, surprisingly, we find these components to be reasonably shared by different layers and heads in a single model, by different models, and across different input distributions (of the same language---English). Based on these insights, we investigate how this structure could be used to reconstruct the full set of attention scores without computing all of them explicitly. We propose an approach for selecting a partial set of scores for exact computation, and for reconstructing all scores from this partial set. Our experiments using this approximate computation approach within transformer models show an encouraging trade-off between network accuracy and attention computation cost.

In summary, our contributions in this paper are as follows:\\
\textbullet\ Through eigen analysis of attention scores of different variants of BERT transformer models~\citep{devlin2018bert}, we establish that these scores lie in low-dimensional subspaces. We show that, surprisingly, these subspaces are largely shared across different layers, models, and datasets.\\
\textbullet\ We propose an approximate attention computation approach that selects a subset of token pairs to compute attention scores exactly, and estimates the remaining from these computed values.\\
\textbullet\ We show that this partial computation approach yields attention score estimates with low mean squared error, and conduct preliminary experiments to train networks featuring these approximations.

\subsection{Background}

\paragraph{Transformers} Each transformer block comprises two components: 1) a multi-head self-attention block; and 2) a token-wise feed-forward multi-layer perceptron (MLP). The input to these blocks is a sequence of vectors $\mX \in \R^{f \times n}$, where $n$ is the sequence length, and the columns of $\mX$ represent the $f$-dimensional embedding of different tokens. The self-attention layer updates these embeddings by a linear combination of values using the pairwise dot product similarities of per token query and key vectors, where values, queries, and keys are all computed by linear transforms applied to $\mX$. The attention scores $\mA_{\mX}$ are computed as:
\begin{equation}
    \mA_{\mX} = \mX^\top \mW_Q^\top \mW_K \mX / \sqrt{d},
\end{equation}
where $\mW$ are trainable parameter matrices, and $\mW_Q \mX$ and $\mW_K \mX \in \R^{d\times n}$ denote the $d-$ dimensional query and key projections. These attention scores are used to linearly combine inputs as follows: $\mZ= \sigma  \left(\mA_{\mX} \right) \cdot \mX^\top \mW_V^\top \cdot \mW_0$, where $\mW_{V}$ denotes the value projection, and $\sigma$ is a row-wise softmax operator. Multi-head attention involves multiple such trainable attention heads in a single layer---using dimensionality $d$ of the queries, keys, and values in each head, being equal and summing up to $f$. The output of the attention block is fed into a tokenwise feedforward layer: $ \mW_2\, {\Phi}\left( \mW_1 \mZ^\top \right),$ with ${\Phi}$ denoting a non-linear activation. Both the self-attention and MLP blocks employ layer-normalization and residual connections.

\paragraph{Related Work} Given the popularity of transformers, there have been many works on understanding their behavior in natural language tasks. \citet{clark2019does,hewitt2019structural,vig2019analyzing, michel2019sixteen} used several language tasks as probes to understand how the language representation evolves over the layers of transformer models. They demonstrated that different heads specialize in particular linguistic sub-tasks such as parts of speech tagging. These works focused on the ability of attention to capture linguistic knowledge. We encourage the reader to  see \citet{rogers2020primer} for an excellent overview of such analyses.

Speeding up attention computation in transformers using different approximations has also been an important research direction. One popular approach is to perform a sparse computation of attention scores. Several works have explored different sparsity patterns and shown their utility for long sequence lengths~\citep{child2019generating, beltagy2020longformer, guo2019star, yun2020}. Others have explored using clustering, hashing, etc.\ to group tokens and compute attention based on these groupings~\citep{kitaev2020reformer, roy2020efficient}. \citep{choromanski2020rethinking} explored using a low rank and kernel approximation of attention scores. We refer to \citet{tay2020long} for a more detailed discussion of these approaches. However, these works are not based on analysis of the distribution of attention scores on inputs from natural datasets. In contrast, our work is primarily focused on analyzing the variation of attention on real-world datasets, which can serve as a starting point for approximation approaches (including the one we propose).

Recently, \citet{raganato2020fixed} explored using fixed attention patterns with different patterns in each head, and one learnable attention head per layer. \citet{tay2020synthesizer} propose using a learnable, but input independent, attention matrix and evaluate such models on language tasks. While these works are similar in spirit they do not analyse attention scores learned by transformers, as we do in this paper. Moreover we keep attention computation input dependent, and exploit the low dimensional structure for reconstruction from partial computation.

\section{Eigen Analysis of the Attention Scores} \label{sec:eigen_analysis}
\insertFigBBGlobalEigen
\insertFigBBEigenAll

In this section we present our analysis of attention scores on a pre-trained $\BB$ model~\citep{devlin2018bert}\footnote{Please see Appendix for a similar analysis of the $\BL$ model.}. This model has 12 transformer layers and is pre-trained using a Masked Language Modeling (MLM) task on English Wikipedia and Books datasets~\citep{zhu2015aligning}. We follow the same setting as in \citet{devlin2018bert} and use their codebase\footnote{https://github.com/google-research/bert.}. We use an input sequence length of 128 for our experiments\footnote{Note that BERT uses a sequence length of 512. However the attention scores covariance matrix in that case has $512^2 \times 512^2 \approx 68B$ entries making it challenging to compute in practice.}. We refer to Appendix for our detailed experimental setup.

Our goal is to analyse the principal components of the subspace that captures the variation of attention scores. Towards this we first compute the covariance matrix of the attention scores for the entire network across all inputs. Let $a_{\mX}^{l ,h} = \text{Vectorize}(\mA_{\mX}^{l ,h})$ be the vectorized form of attention scores from layer $l$ and head $h$ for a given input $\mX$. For a given input sequence length of $n$,  $a_{\mX}^{l ,h} \in \R^{n^2}$. The covariance matrix of the attention scores is given below\footnote{Note that we use the outer-product matrix without mean subtraction, since we want to analyze the span of the attention scores themselves.}.

\begin{align}
    \cov_a = \E_{\mX} \left[ \frac{1}{ L \cdot H}\sum_{l \in [L], h \in [H]}  a_{\mX}^{l ,h} (a_{\mX}^{l ,h})^\top\right].
\end{align}
Here $L$ denotes the number of layers and $H$ the number of heads per layer. For a given sequence length $n$, $\cov_a$ is a $n^2 \times n^2$ dimensional matrix. In practice, we estimate $\cov_a$ by computing an empirical average over all the training examples. We compute this for the Wikipedia dataset by averaging over 2,500M words. 

Let the eigen decomposition of $\cov_a$ be $\sum_{i} \lambda_i v_i v_i^\top$, $v_i \in \R^{n^2}$. Eigenvalues capture the variation of the attention scores distribution along different principal components. We plot the top $100$ eigenvalues of this matrix in Fig.~\ref{fig:eigen_bb_global}. We also plot the cumulative sum of eigenvalues ($\sum_{i=1}^k \lambda_i / \sum \lambda_i$), referred to as cumulative eigen energy in the middle of Fig.~\ref{fig:eigen_bb_global}. We first observe that attention scores lie in an approximately low rank subspace with few eigenvalues dominating over the rest. Even though attention scores are represented in a $16384$ dimensional space, top $200$ ($1.2\%$) eigenvalues capture $>90\%$ of the total energy. Note that this is different from the rank of the attention scores matrix ($\text{rank}(\mA_{\mX})$), and instead captures the subspace dimension of attention scores across different inputs.

One may wonder if this low dimensional nature of attention scores is due to some constraint in the transformer architecture. We note that transformers have $2 \cdot n \cdot d$ degrees of freedom in the query and key projections used for attention computation per layer. For $\BB$ model this translates to $2 \cdot 128 \cdot 768$ degrees of freedom, much larger than the low rank we observed in Fig.~\ref{fig:eigen_bb_global}. Additionally we will see that at initialization the spectrum is quite flat, and model learns this low dimensional structure during training.

\paragraph{Individual layers}
We next look at covariance matrix of attention scores from individual layers.
\begin{align}
    \cov_a^l = \E_{\mX} \left[ \frac{1}{ H} \sum_{\mX \in \gX, h \in [H]}  a_{\mX}^{l ,h} (a_{\mX}^{l ,h})^\top \right].
\end{align}
Note that the global covariance matrix ($\cov_a$) is the mean of individual layer covariances ($\cov_a^l$) across all layers. We plot the cumulative eigen spectrum for layers 0, 5 and 11 of a $\BB$ model in Fig.~\ref{fig:eigen_bb_global}. Again we notice a similar low rank\footnote{Note that we will drop the "approximate" qualifier in the remainder of the paper and simply refer to such matrices as low rank.} structure even in the individual layer covariance matrices. However we do notice difference between layers, with earlier layers having a flatter spectrum than later layers. This can potentially be attributed to earlier layers of the model being more input sensitive than the later layers.

\paragraph{Training steps}
We next plot the evolution of the eigen spectrum at different steps during the training in Fig.~\ref{fig:eigen_bb_ablation}. We first notice that initially the spectrum is quite flat with top $100$ eigenvalues capturing only $35\%$ of the total energy. It however quickly concentrates in $1/4$th of the total training steps, leading finally to an approximate low rank structure with almost $90\%$ energy captured by top $200$ eigenvalues.

\paragraph{Sequence length}
We next study the effect of the input sequence length on the eigen values of the covariance matrix. We consider four different values of input sequence lengths and plot the eigen spectrum in Fig.~\ref{fig:eigen_bb_ablation}. We notice that as we increase the sequence length the rank of the covariance matrix remains relatively small even though the dimension increases quadratically.

\paragraph{Model size}
To study the effect of model size on the eigen spectrum we consider two additional models $\BS$ and $\BL$ with 6 and 24 layers respectively. We plot their eigenspectrum in Fig.~\ref{fig:eigen_bb_ablation}. We notice that though the rank of the covariance matrix increases with model size, it is still relatively small. Even for a large model ($\BL$) top 200 eigen values capture greater than $85\%$ of the total energy.

\paragraph{Dataset}
Finally we compute attention score distributions from different datasets and plot their eigen spectrum in Fig.~\ref{fig:eigen_bb_datasets}. We notice that the same low rank behavior holds across different datasets.

\subsection{Subspace Similarity}
\insertFigBBEigenCross
\insertFigBBDatasets
We next study the similarity between the principal components of the global and different layer covariance matrices. To measure this we project different layer covariance matrices onto the eigenspace of the global covariance matrix. Let $\mV \in \R^{n^2 \times k}$ be the projection matrix onto a $k$ dimensional subspace. Then for a given attention score $a_{\mX}$, we are interested in the following projection norm.
\begin{align}\label{eq:projection}
\E_{\mX} \left[ \frac{1}{L \cdot H} \sum_{l, h}\|\mV^\top a_{\mX}^{l ,h}\|^2 \right]= \E_{\mX} \left[\frac{1}{ L \cdot H} \sum_{ l, h} \Tr(\mV^\top a_{\mX}^{l ,h} (a_{\mX}^{l ,h})^\top \mV )\right] = \Tr(\mV^\top \cov_a \mV).    
\end{align}
Here $\Tr$ computes the trace of a matrix. Note that if $\mV$ spans the top-$k$ eigenspace of $\cov_a$, then this is exactly the sum of its top-k eigen values. The above projection measures how much the principal components $\mV$ capture the variation in attention scores $\cov_a$. 

In Fig.~\ref{fig:eigen_bb_cross} we plot this projection norm of the per-layer covariance matrices ($\cov_a^l$) of a $\BB$ model with the top-$k$ eigen space of $\cov_a$ referred to as {\em Global} in the plot. We also plot the exact eigenspectrum of $\cov_a^l$ for comparison. We notice that projection onto the eigen space of $\cov_a$ preserves most of the eigen spectrum of $\cov_a^l$, showing that their principal components are very similar. We repeat this analysis using the covariance matrices computed at different steps of the training, and notice in Fig.~\ref{fig:eigen_bb_cross} that by $1/3$rd into training, the eigenspace is highly similar to that of the fully trained model. We also compare eigenspace across different sized models and notice that there is a substantial overlap, with top 250 eigen vectors of $\BB$ capturing more than $70\%$ of energy of a $\BL$ model. Finally we compare the subspace similarity of attention scores from different datasets in Fig.~\ref{fig:eigen_bb_datasets} and notice they have high overlap as well.

\insertFigBBGlobalEigenTwoDVisualize

\subsection{Per-Query Attention Scores}
\insertFigBBGlobalEigenOneD
\insertFigBBGlobalEigenOneDVisualize

So far we have analysed the attention scores computed for the entire input. However in transformers, attention is computed independently for each token/query in the input. We now study the eigen spectrum of per-query attention scores which are the rows of the attention scores matrix. 

We first define the per-query attention scores covariance matrices as follows. Let $\mA_{\mX}^{l ,h}[i, :]$ denote the $i$th row of $\mA_{\mX}^{l ,h}$. Then $a_{\mX}^{l ,h, i} = \mA_{\mX}^{l ,h}[i, :]$ is the $i$th query attention score from layer $l$ and head $h$ for a given input $\mX$. For a give input sequence length of $n$,  $a_{\mX}^{l ,h, i} \in \R^{n}$. The covariance matrix of the per-query attention scores, a $n \times n$ matrix, for row $i \in [n]$ is given below.
\begin{align}
    \qcov_a^i = \E_{\mX} \left[ \frac{1}{L \cdot H} \sum_{l, h} a_{\mX}^{l ,h, i} (a_{\mX}^{l ,h, i})^\top.\right]
\end{align}

We plot the eigenvalues of this matrix in Fig.~\ref{fig:eigen_bb_global_1dim} for different input queries/rows, and the fraction of total eigenvalue sum for different queries. We notice that special tokens 0 and 127, which correspond to [CLS] and [SEP] tokens in BERT, have higher rank and lower fraction of energy compared to other tokens. We notice that the eigenspectrum of most other queries is similar with them being predominantly concentrated in first eigenvalue.

Note that $\qcov_a^i$ are block diagonals of $\cov_a$, and do not capture cross query variation. To compare how much the variation in individual rows captures the global variation of attention scores we plot the cumulative eigen energy of global attention scores $\cov_a$ and of the per token attention scores aggregated across rows. We notice that per token eigenspectrum captures most of the variation of attention scores.

Finally we visualize the top 5 eigenvectors of each query stacked across rows in Fig.~\ref{fig:eigen_bb_global_1dim_visualize}. Note the similarity with the corresponding eigen vectors from the global attentions scores in Fig.~\ref{fig:eigen_bb_global_2dim_visualize}.

\section{Reconstructing Attention from Partial Computation}%
\label{sec:sparse}

\newcommand{\covs}{{\bm{C}}}
\newcommand{\att}{a}
\newcommand{\covname}{covariance}
\newcommand{\qcomp}{\overline{P}}
\newcommand{\rmtx}{\bm{R}}

In the last section, we saw that attention scores tend to lie in low-dimensional subspaces. This implies that the attention score matrix, for a given input at a given self-attention layer, can be represented with fewer coefficients than the total number of elements in that matrix. We now investigate whether this property can be used to compute attention scores efficiently, without explicit computation of inner products between all query-key pairs in the input sequence.

\subsection{Formulation}

Let $\att \in \R^{l\times 1}$ denote a vector of attention scores and $\covs \in \R^{{l\times l}}$ its corresponding \covname{} matrix. Here, $\att$ corresponds to either a single row of the attention matrix or to its flattened version, with $l$ being number of tokens $n$ or its square $n^{2}$ in each case. We seek to explicitly compute the values of only a partial subset of $k \ll l$ elements of $\att$ with query-key inner products, and then reconstruct the remaining elements from these computed values.

Accordingly, we let $P \subset \{1, 2, \ldots l\}, |P| = k$ denote the indices of the elements of $\att$ to be computed exactly, and $\qcomp = \{1, 2, \ldots l\} \setminus P$ the set of remaining indices. Moreover, we let $\att_{P} \in \R^{{k}\times 1}$ and $\att_{{\qcomp}} \in \R^{{(l-k)\times 1}}$ denote vectors containing the corresponding elements of $\att$. Thus, after explicitly computing $\att_{P}$, we seek to compute an estimate $\hat{\att}_{\qcomp}$ of the remaining attention scores $\att_{\qcomp}$ from $\att_{P}$. We will do so using a linear transform $\rmtx \in \R^{(l-k)\times k}$ as $\hat{\att}_{\qcomp} = \rmtx\,\att_{P}$.

\paragraph{Optimal Reconstruction} Given a choice of $P$, the average of the squared error $\|\att_{\qcomp} - \rmtx\,\att_{P}\|^{2}$ in the estimates $\hat{\att}_{\qcomp}$ is given by trace of the matrix $\covs_{\qcomp|P}$, which is defined as\footnote{Note that these expressions correspond to to the conditionals of multivariate zero-mean Gaussian distributions with covariance $\covs$.}
\begin{equation}\label{eq:eformula}
  \covs_{{\qcomp}|P} = \mathbb{E}_{\mX} (\att_{\qcomp} - \rmtx\,\att_{P}){(\att_{\qcomp} - \rmtx\,\att_{P})}^{\top} = \covs_{\qcomp{}\qcomp}  -\rmtx \covs_{P\qcomp{}} -\covs_{\qcomp{}P}\rmtx^{\top} + \rmtx\covs_{PP}\rmtx^{\top}.
\end{equation}
Here, $\covs_{AB}$ denotes a ``crop'' of the \covname{} matrix $\covs$ containing the rows and columns with indices in sets $A$ and $B$. It is easy to see that the trace of $\covs_{\qcomp{}|P}$ is  minimized by setting $\rmtx$ as
\begin{equation}
  \label{eq:optr}
 \rmtx = \covs_{\qcomp{}P}\covs_{PP}^{{-1}},
\end{equation}
and the expression for $\covs_{\qcomp|P}$ simplifies to the Schur's complement of $\covs_{PP}$ in $\covs$, i.e.,
\begin{equation}
  \label{eq:rescov}
  \covs_{{{\qcomp|P}}} = \covs_{\qcomp{}\qcomp} - \covs_{\qcomp{}P}\covs_{PP}^{{-1}}\covs_{P\qcomp{}}.
\end{equation}

\paragraph{Selecting Partial Set} Our choice of the partial set of indices $P$, for a given choice of its size $k$, should be such that it yields a low reconstruction error (from~\eqref{eq:rescov}). Unfortunately, finding the globally optimal choice of $P$ would require evaluating all possible subsets of size $k$, with computational cost $O(\exp(k))$. But, we find a greedy selection approach as described below to work well in practice.

We form a series of matrices $P^{1}, P^{2}, \ldots P^{k}$, with $|P^{k'}| = k'$ and $P^{k'}\supset P^{k'-1}$ (with $P^{0}$ the empty set), and then set $P = P^{k}$ for our desired choice of $k$. Given $P^{k'}$ and the corresponding residual \covname{} matrix $\covs^{k'} = \covs_{\qcomp^{k'}|P^{k'}}$, we set $P^{k'+1} = P^{k'}\cup \{i\}$, choosing $i \in \qcomp^{k'}$ as
\begin{equation}\label{eq:addindex}
  i = \arg \min_{i} \mbox{Tr}(\covs^{k'+1})
     = \arg \min_{i}\,\, \sum_{j \in \qcomp_{k'}\setminus\{i\}} \covs^{{k'}}_{jj} - \frac{{(\covs^{{k'}}_{ij})}^{2}}{\covs^{k'}_{ii}} 
    = \arg \max_{i}\,\, \frac{\sum_{j \in \qcomp_{k'}} {(\covs^{{k'}}_{ij})}^{2}}{\covs^{k'}_{ii}}.
\end{equation}

\paragraph{Computational Cost} The computational savings of this approach will depend on whether it is applied to the whole attention matrix, or to each row independently. We let $\bar{k}$ denote the total number of scores we compute exactly, with $\bar{k}=\,nk$ and $k$ for the per-query and whole matrix settings. Then, for $d-$dimensional query and key vectors, the combined computational cost of exact computation and reconstruction using our approach is $O(\bar{k}d + \bar{k}n)$ and $O(\bar{k}d + \bar{k}n^{{2}})$ in the per-query and whole matrix settings respectively. In contrast, full exact computation has a cost of $O(n^{2}d)$. Thus, although the whole matrix setting may yield better reconstructions by exploiting correlations across rows for the same number of partial computations $\bar{k}$, it also entails a higher computational cost. As we see next, the per-row setting yields a better trade-off between accuracy and computational cost.


\subsection{Reconstruction Error}

We begin by evaluating the partial pairs $P$ selected by our method for exact computation, and the corresponding average squared reconstruction errors, for attention scores from a typical network. Note that the reconstruction errors can be computed directly (using~\eqref{eq:rescov}) from the \covname{} matrices estimated in Sec.~\ref{sec:eigen_analysis}. We report results for the $\BB$ model trained on the standard pre-training task of masked language modeling (MLM)~\citep{devlin2018bert}, using average covariance across layers and heads.

\begin{figure}[!t]\centering
\includegraphics[width=0.98\textwidth]{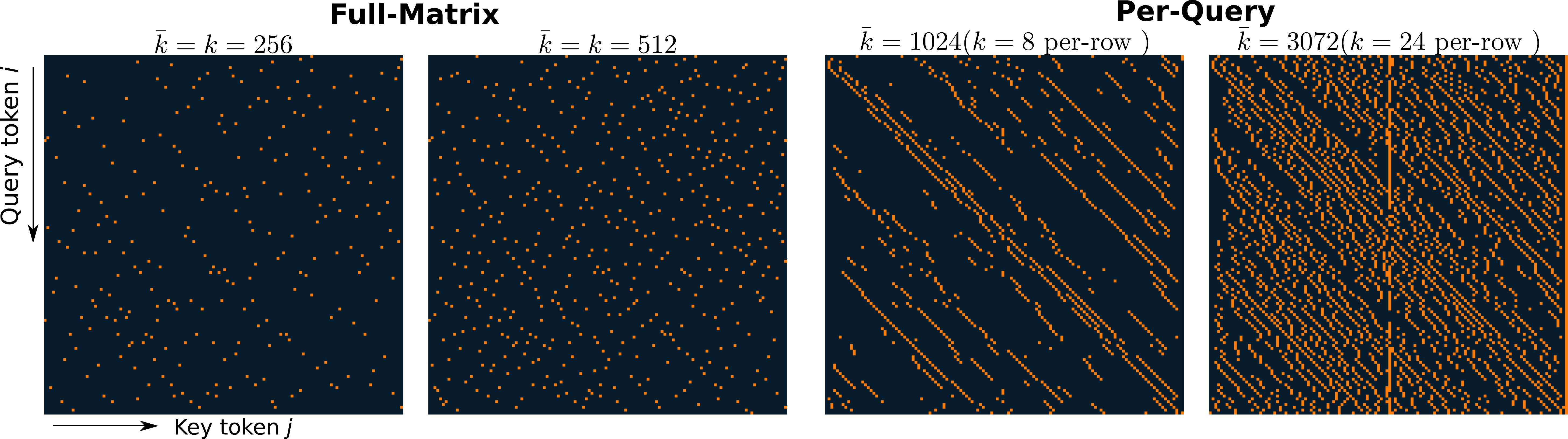}\vspace{-1em}
\caption{\textbf{Pairs Selected for Partial Computation.} We visualize examples of the partial sets $P$ of query-token pairs selected by our greedy algorithm (eq.~\eqref{eq:addindex}), given total number of pairs $\bar{k}$. We show examples of both whole matrix and per-query sets, with $k=\bar{k}/n$ pairs for each row for the latter.}\label{fig:qpatterns}   \vspace{-0.1in}
\end{figure}

\begin{figure}[!t]\centering
\includegraphics[width=0.45\textwidth]{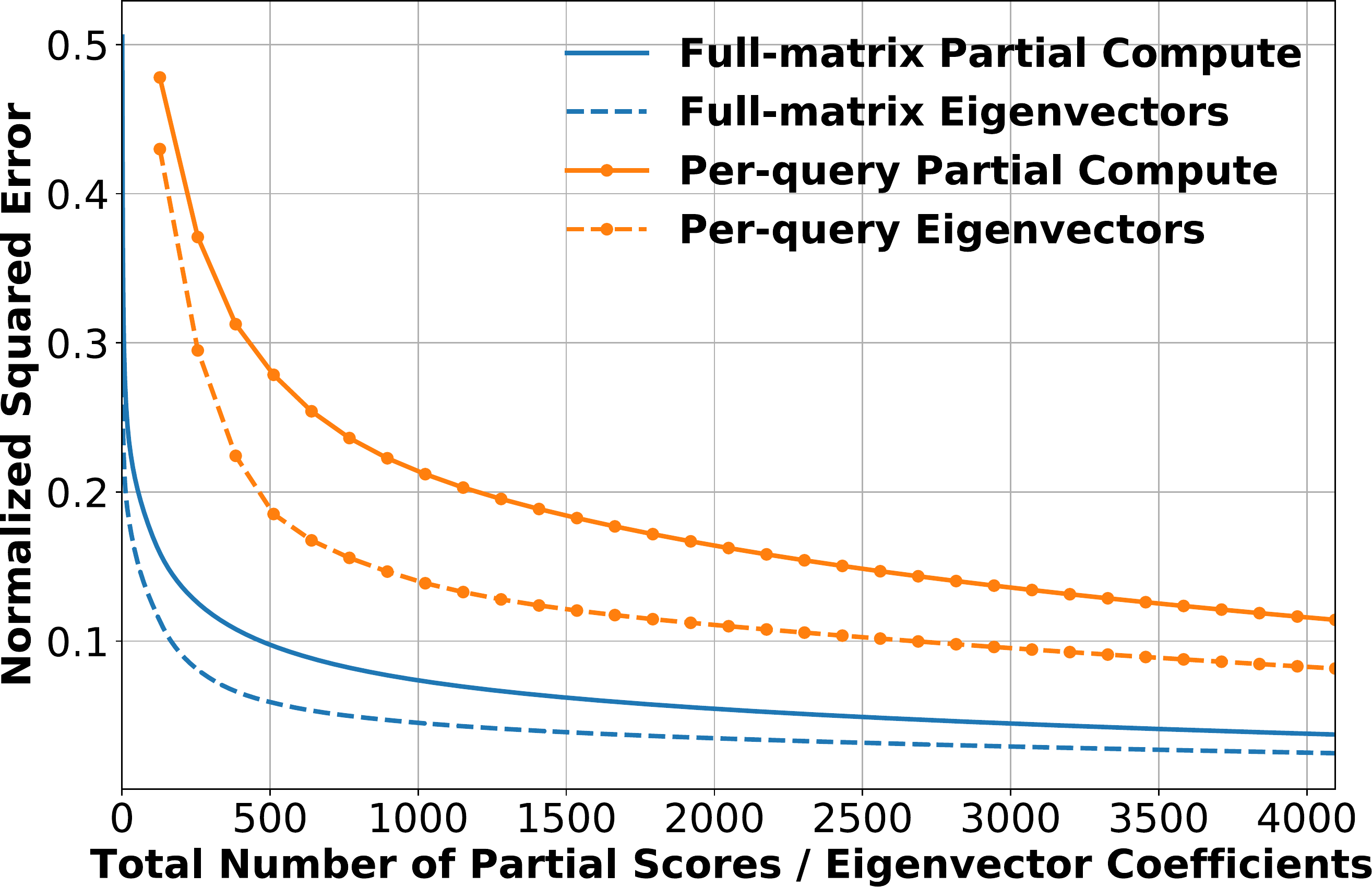}~~~~\includegraphics[width=0.45\textwidth]{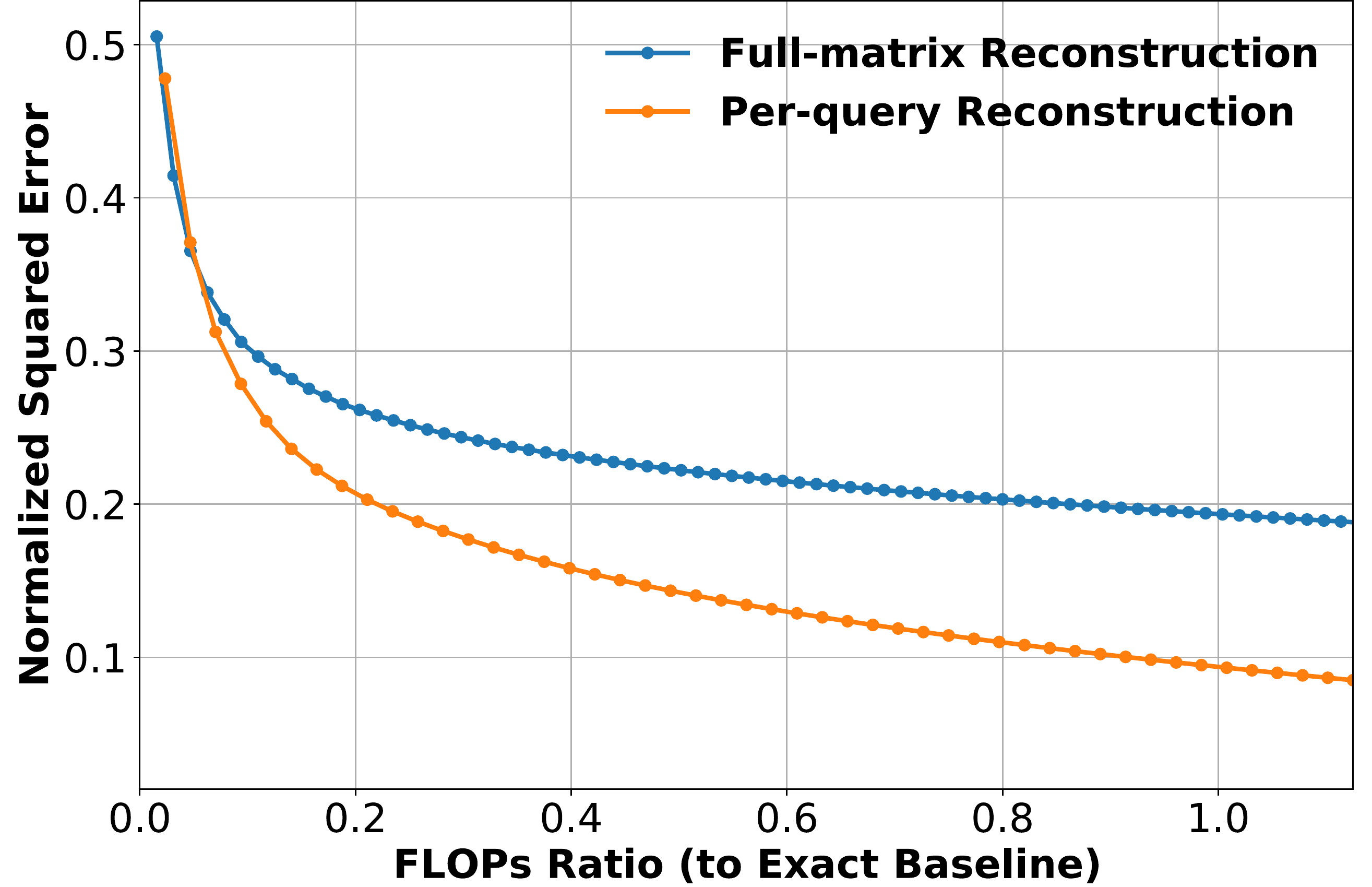}\vspace{-0.5em}
\caption{\textbf{Reconstruction Errors from Partial Computation}. We show average squared error of attention scores---normalized by their total variance---reconstructed using our partial computation approach. (Left) We plot these for both whole matrix and per-query reconstruction, as a function of the total number $\bar{k}$ of pairs computed exactly. We compare these to reconstruction from projections on to an equivalent number of dense eigenvectors. (Right) We analyze the accuracy-computation trade-off in whole matrix vs.\ per-query reconstruction, but plotting errors as a function of ratio of floating point operations needed for partial computation and reconstruction, to those required for full exact computation (FLOPs ratio).}\label{fig:sparse_error}
  \vspace{-0.1in}
\end{figure}

In Fig.~\ref{fig:qpatterns}, we show the query-token pairs selected by our greedy algorithm for a few choices of $k$---for both the whole matrix and per-query settings. We notice that, like the eigenvectors in Sec.\ref{sec:eigen_analysis}, these patterns often cluster along shifted diagonals. This effect is more pronounced in the per-query patterns, where each row can rely only on its own exact computations, while those for the whole matrix are less coherent.
We next characterize the reconstruction error in both settings in Fig.~\ref{fig:sparse_error}. First, we plot reconstruction errors, normalized by total
variance of the full covariance matrix, for a range of values for number of exact scores $\bar{k}$ (equal to $k$ for whole matrix patterns, and $nk$ for per-query). For reference, we compare these to error from approximation by the same number of top whole matrix and per-query eigenvectors. As expected, the constraint on sampling a subset of entries rather than projecting to dense eigenvectors leads to a gap in error. Nevertheless, we find that our approach yields reasonable reconstructions with increasing numbers of coefficients. When comparing with an equivalent number of exact computations, using whole matrix reconstruction yields better results. But Fig.~\ref{fig:sparse_error} also provides a comparison in terms of equivalent computational cost, and we see here that the per-query setting affords a better trade-off due to its lower cost of reconstruction.

\subsection{Network Performance}
We next look beyond squared errors in reconstructed attention scores, and evaluate the effect of this approximation on overall network performance. As expected, simply introducing the approximation in a network that has been trained with exact attention performs poorly (see supplement). Instead, we consider training transformer models with the partial computation and reconstruction built-in.

We evaluate approximation performance in the $\BB$ model, on the pre-training MLM task~\cite{devlin2018bert} followed by fine-tuning for entailment classification on the Multi-Genre NLI corpus~\citep{mnli} (a part of the GLUE benchmark~\citep{wang2019glue}). We introduce approximate attention computation in all but the last layer (where typically only the embedding of a single token is retained, and approximation would offer no computational benefit), and use per-query reconstruction from partial sets of different numbers of exact query-key pair scores $k$. We select these sets $P$ using our greedy algorithm, and then initialize the reconstruction matrix $\bm{R}$ to its optimal value computed using \eqref{eq:optr}---based on a covariance matrix computed from all layers and heads of a baseline exact $\BB$ model.

All weights of the network are trained end-to-end, back-propagating through the partial selection of exact dot products in $P$, and reconstruction with the linear transform $\bm{R}$. Moreover, while the set $P$ is kept constant, we evaluate different approaches for $\bm{R}$. In one approach, we keep $\bm{R}$ fixed to its optimal initialized value. We also consider updating $\bm{R}$ during training---as a common matrix for all layers, as well as learning a different $\bm{R}$ for each layer.

These results are summarized in Table~\ref{tab:sparse_accuracy}. We find that training $\bm{R}$ rather than keeping it fixed is beneficial---suggesting that its initial value, being optimized for squared error, may not be optimal for accuracy. Interestingly, training a common $\bm{R}$ for all layers appears to be beneficial for smaller values of $k$, while larger $k$ benefits from per-layer training. Overall, we find only a modest drop in accuracy for a reduction of 25-45\% in the attention computation cost (for $k=32$ and $24$).
\begin{table}[t]
  \caption{\textbf{Network Performance with Partial Computation}. We train BERT models with partial attention computation and per-row restoration, for different values of per-row exact scores $k$. We report accuracies of models trained for MLM and fine-tuned for MNLI (averaging three runs for the latter), and FLOPs ratios for attention computation. We select the optimal set $P$ and initialize $\bm{R}$ as per~\eqref{eq:optr}, and consider three training regimes---\textbf{(F)} where the $\bm{R}$ matrix is kept fixed to its initialization; \textbf{(C)} where a common $\bm{R}$ is trained for all layers; and \textbf{(P)} where a separate $\bm{R}$ is trained for each layer.}\label{tab:sparse_accuracy}
  \centering
  \begin{tabular}{cccccccccccccc}
    \toprule
    & \bf Exact &~& \multicolumn{3}{c}{\bf $\bm{k=16}$} &~&  \multicolumn{3}{c}{\bf $\bm{k=24}$} &~& \multicolumn{3}{c}{\bf $\bm{k=32}$}\\
    & \bf Baseline &&\bf F &\bf C &\bf P &&\bf F &\bf C &\bf P &&\bf F &\bf C &\bf P \\\midrule

    \multicolumn{6}{l}{\em Test Accuracy}& \\
    MLM & \bf 66.0 && 63.9& \bf 64.5& 63.6&& 63.1& \bf 65.5& 64.7&& 64.5& 64.7& \bf 65.6\\
    MNLI & \bf 81.6 && 75.8 & \bf 77.1 & 76.7 && 76.5 & \bf 79.3 & 78.7 && 77.2 & \bf 79.7 & \bf 79.7\\

    \midrule

    FLOPs & \multirow{2}{*}{1.0} && \multicolumn{3}{c}{\multirow{2}{*}{0.375}} && \multicolumn{3}{c}{\multirow{2}{*}{0.5625}} && \multicolumn{3}{c}{\multirow{2}{*}{0.75}}\\
    Ratio &\\

    \bottomrule
  \end{tabular}
  \vspace{-0.05in}
\end{table}


\section{Discussion}

In this paper, we analyzed the distribution of attention scores generated by transformers on natural language inputs, and found them to lie in a relatively low-dimensional subspace. We found this behavior to hold across different layers and models, and found significant overlap between their eigen subspaces---indicating that this phenomenon is fundamentally a product of the underlying language structure. Our analysis can serve as a useful and principled foundation for approximate attention approaches, and we propose one such approach based on partial computation followed by reconstruction. Our results, both in terms of squared reconstruction error and trained network performance, indicate that this is a promising direction for future research.

 We want to emphasize, however, that our specific partial computation and reconstruction method is only one possible way of exploiting this low-dimensional variance structure, and we expect future work will explore others. Moreover, while our analysis was restricted to English language datasets, a natural question to ask is whether these findings also hold for other languages, and to other domains---such as computer vision, where transformers act on tokens representing image patches instead of words. Finally, a limitation of our method is that it assumes a fixed sequence length and is challenging to employ when this length is large. Generalizing our approach to work on longer sequences---potentially by modeling correlations within sub-sequences and applying it on all translated sub-sequences---is another interesting direction of future work.

\bibliographystyle{plainnat}
\begin{small}

\end{small}


\if11
\clearpage
\appendix

\pagenumbering{roman}
\begin{figure*}[!t]
\begin{center}
{\huge \bf Supplementary Material}
\end{center}
\end{figure*}
\setcounter{figure}{9}

\section{Network Training Details}\label{sec:experiment_setup}
\begin{table}[!t]
\caption{Details of the BERT models used in this paper.}
\label{table:bert}
\centering
\begin{tabular}{c c  c c} 
 \toprule
 \bf Model  & {\rm Layers} & {\rm Hidden size} & {\rm Num heads} \\ [0.5ex] 
 \midrule
  $\BS$ & 6 & 768  & 12 \\
$\BB$ & 12 & 768  & 12 \\
$\BL$ & 24 & 1024  & 16 \\
 \bottomrule
\end{tabular}
\end{table}

We used the same setting as in BERT~\citep{devlin2018bert}, including using their codebase\footnote{https://github.com/google-research/bert}, to train the various Transformer models. We pre-trained the models on English Wikipedia and Books datasets \citep{zhu2015aligning}. We used inputs of sequence length 128 and trained the model using the Masked Language Modeling (MLM) and Next Sentence Prediction (NSP) tasks. We trained these models for 450k steps with a batch size of 1024, using the Adam optimizer with a peak learning rate of 1e-4, a linear warmup for the first 10k steps, followed by linear decay. We also used weight decay of 1e-4 and dropout of 0.1.

\section{Eigen analysis of $\BL$ model}\label{sec:bert_large}

Figures \ref{fig:eigen_bl_global}-\ref{fig:eigen_bl_global_1dim_visualize} present eigen analysis results analogous to those shown in Sec.~\ref{sec:eigen_analysis} for a $\BL$ model. We notice similar behavior as the $\BB$ model with approximate low rank attention scores variation and large subspace overlap across different settings.

\insertFigBLGlobalEigen
\insertFigBLEigenAll
\insertFigBLEigenCross
\insertFigBLGlobalEigenOneD
\insertFigBLGlobalEigenTwoDVisualize
\insertFigBLGlobalEigenOneDVisualize
\section{Mean-subtracted Attention Scores}

All of our analysis in Sec.~\ref{sec:eigen_analysis} was on the variability of raw pre-softmax attention scores. Although the softmax operation is invariant to the mean value of each row in the attention score matrix, we did not subtract this mean in our main analysis in order to characterize the variability in the raw query-key similarities, and because during partial computation, the value of this mean would be unknown.

\begin{figure}[!t]
\centering
\includegraphics[width=0.5\textwidth]{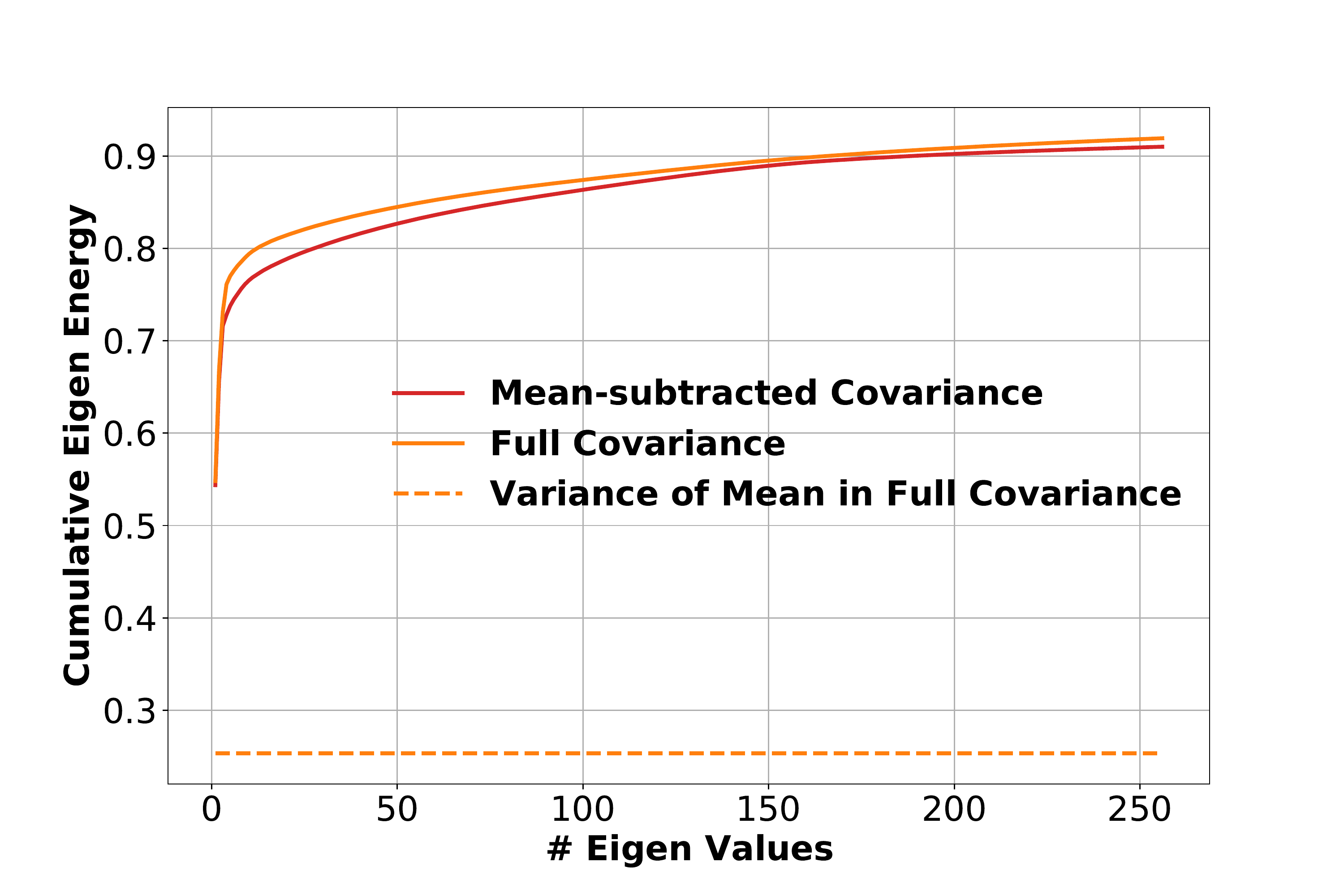}
\caption{\textbf{Analysis after Per-row Mean Subtraction.} Here, we compare the eigen value energies for a covariance matrix from the attention scores of a $\BB$ model, to a version of the covariance matrix computed after subtracting the mean score value in each row of each sample. We find that both versions of the covariance matrix exhibit similar low rank behavior, and that the energy of per-row means makes a negligible contribution to the overall variability of attention scores.}\label{fig:msub}
\end{figure}

For completeness, we also present the eigenspectrum of covariance matrices where all per-row means have been removed in Fig.~\ref{fig:msub}, showing the fraction of total energy captured by the principal eigenvectors (here, both total energy and eigenvectors and values are computed from the modified covariance matrix). We compare this to the energy profile of the original covariance matrix, and find them to be largely similar. We also plot the fraction of energy contribution from variability of the per-row means in the original covariance matrix, and find it to be roughly half of the first eigenvalue. Thus, inclusion of per-row means in our analysis has no meaningful effect on the conclusions.

\section{Approximate Attention during Inference}

In Table~\ref{tab:sparse_accuracy}, we considered the effect on accuracy from using approximate attention scores (due to reconstruction from partial computation), with models that were trained with this approximation. In Table~\ref{tab:1d_projection}, we report the effect of taking a standard model trained with exact attention scores and introducing the use of approximate attention only during inference. We show results for using reconstructions from partial computation, as well as from replacing the exact attention scores in each row with their best approximation from a limited number of per-query eigenvectors. We find that the performance in this case is notably worse than in Table~\ref{tab:sparse_accuracy} where, unlike in this case, all layers in the models had the opportunity to adapt to the attention approximation.

\begin{table}[t]
  \caption{\textbf{Performance of Pre-trained Network with Approximate Attention during Inference}. We evaluate the effect of using approximate attention scores during inference with  a standard $\BB$ model trained with exact attention computation. We report results for two approaches to attention approximation: \textbf{EP:} which approximates attention scores using projections onto the per-query top-$k$ eigenvectors, and \textbf{PC:} where scores are reconstructed from $k$ exact attention scores per query.}
  \label{tab:1d_projection}
  \centering
  \begin{tabular}{cccccccccccccc}
    \toprule
    & \bf Exact &~& \multicolumn{2}{c}{\bf $\bm{k=16}$} &~&  \multicolumn{2}{c}{\bf $\bm{k=32}$} &~& \multicolumn{2}{c}{\bf $\bm{k=64}$}\\
    & \bf Baseline &&\bf EP &\bf PC &&\bf EP &\bf PC &&\bf EP &\bf PC \\\midrule
    MLM Accuracy & \bf 66.0 && 50.8& 51.6 && 55.5& 54.9 && 61.0& 59.7\\

    \bottomrule
  \end{tabular}
  \vspace{-0.05in}
\end{table}
\fi
\end{document}